\documentclass[acmsmall,screen]{acmart}
\usepackage{comment}
\usepackage{lineno}
\usepackage{enumitem}
\usepackage{url}
\usepackage{graphics}
\usepackage{tikz}
\usepackage{caption}
\usepackage{multirow}
\usepackage{graphicx}
\usepackage{enumitem}
\usepackage{tabularx}
\usepackage{multirow}
\usepackage{amsthm}
\usepackage{listings}
\usepackage{float}
\usepackage{wrapfig}
\usepackage{comment}
\usepackage{xcolor}
\usepackage{fontsize}
\usepackage{subfigure}
\usepackage[normalem]{ulem}
\usepackage[ruled,lined,boxed,linesnumbered]{algorithm2e}
\SetKwInput{KwOutput}{Output}
\SetKwInput{KwInput}{Input}
\newcommand{\nosemic}{\renewcommand{\@endalgocfline}{\relax}}
\newcommand{\dosemic}{\renewcommand{\@endalgocfline}{\algocf@endline}}
\let\oldnl\nl
\newcommand{\nonl}{\renewcommand{\nl}{\let\nl\oldnl}}

\definecolor{mypurple}{RGB}{128,0,128}
\definecolor{mybrown}{RGB}{165,42,42}
\newcommand{\sk}[1]{{\color{black}#1}}

\newcommand{\customsize}{\fontsize{7.5}{11}\selectfont} 

\newcommand{\tcf}{f_{\mbox{\scriptsize \em target}}}
\newcommand{\ecf}{f_{\mbox{\scriptsize \em effective}}}

\begin{document}

\title{An Open-Source ML-Based Full-Stack Optimization Framework for Machine 
Learning Accelerators}
\author{Hadi Esmaeilzadeh}
\email{hadi@eng.ucsd.edu}
\orcid{0000-0002-8548-1039}
\author{Soroush Ghodrati}
\email{soghodra@ucsd.edu}
\orcid{0000-0001-5514-8027}
\author{Andrew B. Kahng}
\email{abk@ucsd.edu}
\orcid{0000-0002-4490-5018}
\author{Joon Kyung Kim}
\email{jkkim@eng.ucsd.edu}
\orcid{0000-0003-2698-7950}
\author{Sean Kinzer}
\email{skinzer@eng.ucsd.edu}
\orcid{0000-0002-0955-585X}
\author{Sayak Kundu}
\email{sakundu@ucsd.edu}
\orcid{0000-0002-8077-1328}
\author{Rohan Mahapatra}
\email{rohan@ucsd.edu}
\orcid{0000-0002-2887-9761}
\affiliation{
    \institution{University of California, San Diego}
    \city{La Jolla}
    \state{California}
    \country{USA}
}
\author{Susmita Dey Manasi}
\email{manas018@umn.edu}
\orcid{0000-0001-9358-6255}
\author{Sachin S. Sapatnekar}
\email{sachin@umn.edu}
\orcid{0000-0002-5353-2364}
\affiliation{
    \institution{University of Minnesota}
    \city{Minneapolis}
    \state{Minnesota}
    \country{USA}
}
\author{Zhiang Wang}
\email{zhw033@ucsd.edu}
\orcid{0000-0002-6669-9702}
\affiliation{
    \institution{University of California, San Diego}
    \city{La Jolla}
    \state{California}
    \country{USA}
}
\author{Ziqing Zeng}
\email{zeng0083@umn.edu}
\orcid{0000-0002-6981-2299}
\affiliation{
    \institution{University of Minnesota}
    \city{Minneapolis}
    \state{Minnesota}
    \country{USA}
}

\renewcommand{\shortauthors}{Esmaeilzadeh et al.}

\begin{abstract}
  Parameterizable machine learning~(ML) accelerators are the
  product of recent breakthroughs in ML. To fully enable their
  design space exploration~(DSE), we propose a 
  physical-design-driven, learning-based prediction
  framework for hardware-accelerated deep neural network~(DNN)
  and non-DNN ML algorithms. It adopts a unified approach that
  combines backend power, performance, and area (PPA) analysis
  with frontend performance simulation, thereby achieving a
  realistic estimation of both backend PPA and system metrics
  such as runtime and energy. \sk{In addition, our framework
  includes a fully automated DSE technique, which optimizes
  backend and system metrics through an automated search of
  architectural and backend parameters. Experimental studies
  show that our approach consistently predicts backend PPA
  and system metrics with an average 7\% or less prediction
  error for the ASIC implementation of two deep learning
  accelerator platforms, VTA and VeriGOOD-ML, in both a
  commercial 12 nm process and a research-oriented 45 nm
  process.}
\end{abstract}




\maketitle
\section{Introduction}
\sk{Recent advances in machine learning (ML) algorithms have catalyzed
an increasing demand for application-specific ML hardware accelerators.}
The design of these accelerators is non-trivial: the design cycle
from architecture to silicon implementation often takes months to
years and involves a large team of cross-disciplinary experts.
When faced with stringent time-to-market requirements,
it is imperative to reduce turnaround time without sacrificing 
product quality.

\sk{Recent research has developed automation flows 
for generating parameterizable accelerators,
suitable for both FPGA and ASIC platforms.}
Parameterizable deep neural networks (DNNs) accelerators
include VTA~\cite{VTA, IntelVTA}, GeneSys~\cite{RTML}, and
Gemmini~\cite{Gemmini}. Accelerators for non-DNN ML
algorithms~\cite{Tabanelli21},  such as support vector
machines or linear/logistic regression, have widespread
applications, but have seen more limited research, with
the TABLA platform~\cite{TABLA} being a prominent example
of a general-purpose non-DNN accelerator.

Generators such as those listed above allow designers
to configure key parameters of DNN/non-DNN ML
accelerators, e.g., the number of processing units
or the on-chip memory configuration. The accelerator
hardware description is then automatically translated
to hardware at the register-transfer level (RTL).
The search for an optimal configuration involves
tradeoffs between the {\em power dissipation},
{\em performance}, and {\em area} (PPA) of the
hardware platform, and the {\em energy} and
{\em runtime} required to execute an ML algorithm
on the platform. Therefore, this optimization
involves the solution of two problems:
(i)~generating an ML accelerator that optimizes
the PPA metrics; and
(ii)~selecting a PPA-optimized hardware
configuration that optimizes system-level
metrics such as the runtime and energy
required to run an ML algorithm. Our work
is the first to solve these two problems.

\begin{figure}
    \centering
    \begin{subfigure}[]
        {\includegraphics[width=0.65\columnwidth]{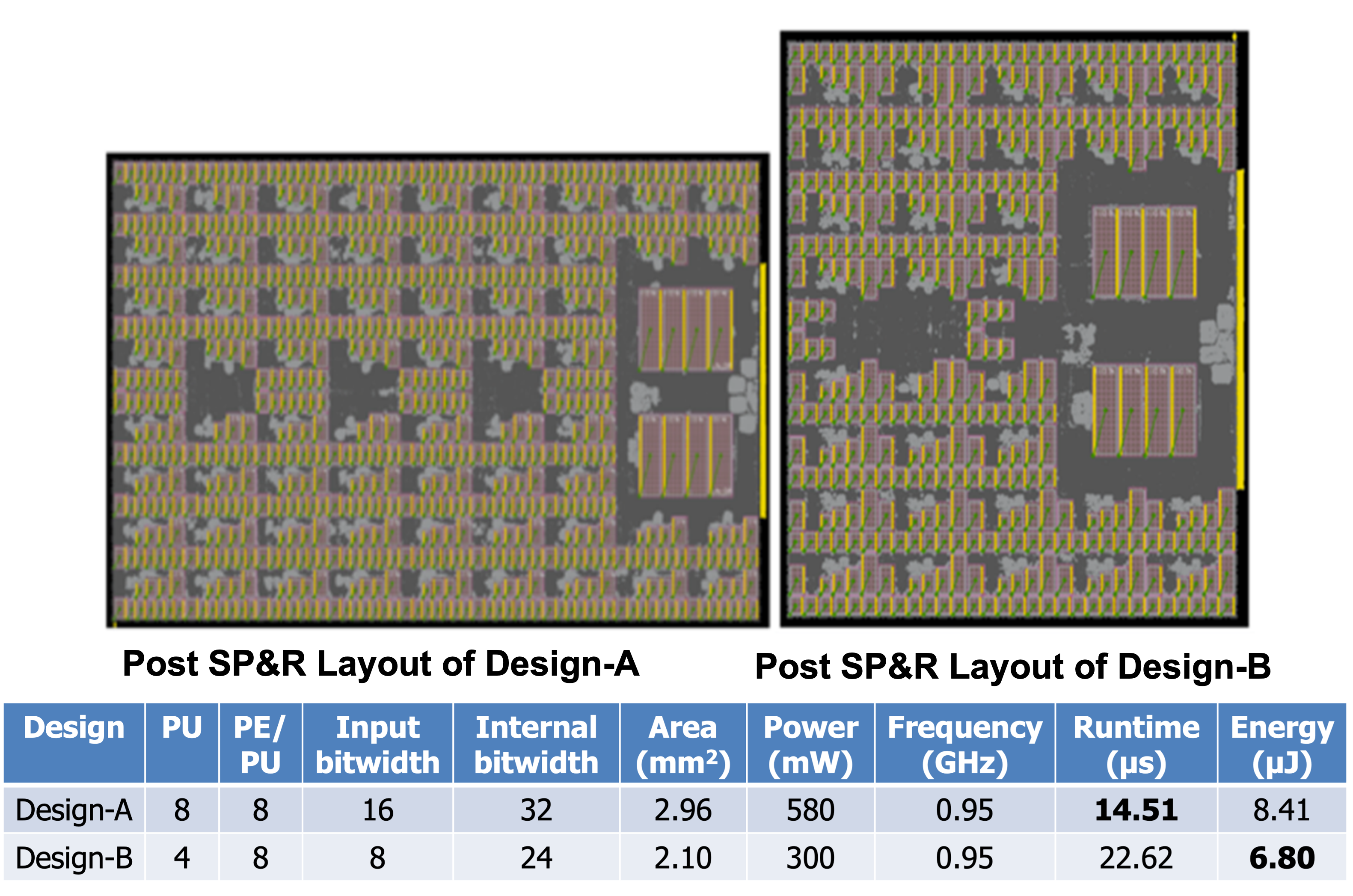}}
    \end{subfigure}
    \begin{subfigure}[]
        {\includegraphics[width=0.65\columnwidth]{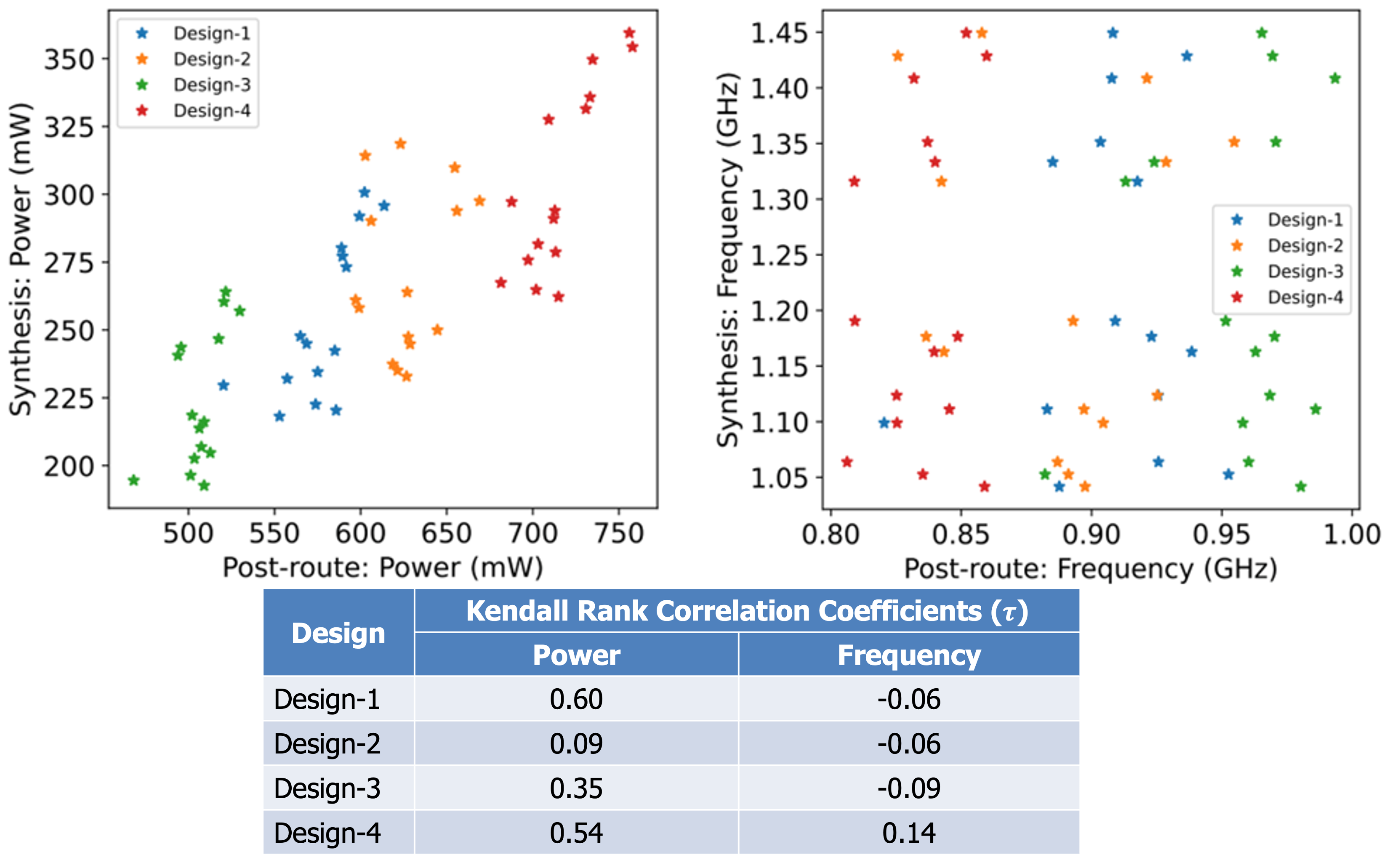}}
    \end{subfigure}
    \caption{ 
    \sk{(a) Post-SP\&R layouts of two TABLA designs implementing 
    the same ML algorithm. Design-A achieves better runtime,
    while Design-B achieves better energy.}
    (b) Miscorrelation of post-synthesis and post-routing
    metrics: total power and clock frequency for TABLA designs.}
    \label{fig:miscorrelation}
\end{figure}

The prediction of {\em platform PPA} based on
an architectural description is a longstanding
challenge in electronic design automation. In
modern nanoscale technologies, the link between
physical design and PPA is particularly acute.
Moreover, for many ML hardware platforms,
a considerable fraction of the layout area is
occupied by large memory macros whose presence exacerbates
the problem of PPA prediction. The prediction
of {\em system-level metrics}, such as the
runtime and energy required to execute an ML
algorithm, is performed using system-level
simulation engines. These simulators model
data transfer and computation within the
accelerator to determine the number of
operations, stall cycles, memory latencies,
etc. Since they use the frequency and power
metrics of the hardware platform as inputs,
their accuracy depends on the quality of
PPA prediction.

Given an ML accelerator, a target clock period,
and a target floorplan utilization, the metrics
of interest are the PPA of the hardware, and the
energy and runtime required to execute ML
algorithms. However, optimizing PPA does not
necessarily guarantee optimal runtime and energy
consumption. 
\sk{For instance, smaller hardware may consume 
less power but may not necessarily deliver the
required improvements in energy consumption or
meet runtime criteria. In 
Figure~\ref{fig:miscorrelation}(a), 
which shows two designs implementing the same ML algorithm,
we see that
Design-B achieves 48\% better power efficiency
than Design-A. However, in terms of energy
efficiency, it only shows 20\% improvement
and does so at a significantly slower runtime.}
This illustrates why Design Space Exploration (DSE)
necessitates a rapid evaluator
capable of assessing PPA, runtime, and energy
consumption for numerous architectural
configurations within a given design
space. A DNN accelerator
may easily have 5-10 million instances, and
conventional evaluators require several days
of synthesis, place and route (SP\&R) runs
to evaluate even a single configuration.
Parallel evaluation runs cannot offer relief
due to limited compute resources and
available EDA tool licenses. Using
post-synthesis PPA without P\&R is inadequate:
Figure~\ref{fig:miscorrelation}(b)~shows poor
correlation between the post-synthesis and
post-SP\&R results for TABLA designs, visually
and through the Kendall rank correlation
coefficient ($\tau$), \sk{where 0
signifies no correlation and \textpm 1
indicates strong correlation or anti-correlation}.
For example, the $\tau$ values of four TABLA
designs for total power are $0.61$, $-0.20$,
$0.07$, $0.47$, and for effective clock frequency
are $0.45$, $-0.20$, $-0.16$, $0.10$.

\sk{
To fully harness the potential of parameterizable
ML accelerators, we propose a physical-design
driven, learning-based prediction framework for
hardware-accelerated ML algorithms. Our ML-based
method accurately predicts PPA and system
performance, and overcomes the
limitations associated with the high
computational expense of design space
exploration. We accomplish this by using a
manageable number of SP\&R backend runs to train
ML models, which in turn predict the performance
of ML accelerator designs for unseen configurations.
Furthermore, we incorporate Multi-Objective 
Tree-structured Parzen Estimator (MOTPE)-based
Bayesian optimization to automatically
optimize the ML
accelerators for the target ML algorithm and metric
requirements, using the trained models. Whereas previous
works primarily focus on architectural design
and/or RTL generation, our framework offers a
full-stack optimization solution for ML accelerators,
encompassing all aspects, from high-level ML algorithm
specification to SP\&R implementation guidelines.
The main contributions of our work are as follows.
}

\noindent
\begin{itemize}[noitemsep,topsep=0pt,leftmargin=*]
\item 
\sk{We propose a machine learning-based full-stack
optimization framework for ML accelerators. This
framework covers key design components within the
software-hardware stack, which include target ML
algorithms, architectural parameters, RTL
generation, SP\&R recipes for hardware
implementation, performance simulation and
design space exploration. As far as we know,
our work is the first of its
kind\footnote{A preliminary version of this
work was published in~\cite{Esmaeilzadeh22}.}
to integrate backend SP\&R recipes into a
framework for optimizing machine learning
accelerators. These implementation recipes are
essential, as they make it considerably easier
to implement the ML accelerator and achieve
the anticipated performance.}

\item \sk{
We build upon our previous 
work~\cite{Esmaeilzadeh22} to include the
prediction of the chip area of ML
accelerators. Furthermore, we
incorporate target floorplan
utilization as a backend knob feature,
enabling the prediction of backend PPA
and system-level performance across
a range of target floorplan utilizations.
}
\item \sk{
We explore three different sampling 
methods: Latin Hypercube sampling (LHS),
and two forms of low-discrepancy sequence
(LDS)-based sampling, specifically using Sobol
and Halton sequences, across various sample
sizes. The choice of efficient sampling
method and appropriate sample size
is pivotal in constructing high-accuracy
prediction models. Additionally, the
generation of testing datasets using these
sampling methods ensures broad coverage
of the design space, thereby enhancing
the reliability of the ML model for DSE.
}
\item 
\sk{
We introduce a physical-design-driven,
learning-based prediction methodology
and a MOTPE-based method to automatically
optimize ML accelerators for given
target ML algorithm and metric
requirements. Our experimental results
suggest that our method can significantly
reduce the implementation time of optimized
machine learning accelerators, reducing
both human effort and tool runtime from
months to days.
}
\item 
\sk{
We present a novel method that capitalizes
on the high degree of modularity present
in machine learning accelerators. It
generates a logical hierarchy graph,
in which each leaf node represents a
building block of the ML accelerators.
This methodology employs a Graph
Convolutional Network (GCN) to extract
graph embeddings and train the model.
Our experiment results indicate
that the GCN model, even when trained on
less data, can match or even outperform
other models in predicting the test dataset.
}
\end{itemize}

\sk{
In addition to the new methodologies 
proposed herein, this work improves 
from our previous 
work~\cite{Esmaeilzadeh22} in several
significant ways: (i) we employ LHS
generated testing dataset, which covers
the entirety of the design space, as
opposed to randomly selecting
configurations from the complete
dataset; (ii) we leverage MOTPE-based
automated DSE, in contrast
to the ``brute-force'' approach used
previously; and (iii) we have expanded
our study to include the open-source
NanGate45~\cite{ng45} enablement and have made
our flow scripts, model training and
DSE code publicly accessible in
the VeriGOOD-ML GitHub
repository~\cite{veriGoodMLRepo}.
}

\sk{
The rest of this paper is organized
as follows. Section~\ref{sec:related_work}
reviews the relevant literature on PPA
prediction. Section~\ref{sec:notations}
explains the key definitions and notations
used in this paper. 
Section~\ref{sec:problem_formulation}
presents the problem formulation for
PPA prediction and design space exploration
for ML accelerators. Section~\ref{sec:approach}
provides an overview of our approach, with brief
descriptions of each component of our proposed
framework. Section~\ref{sec:graph_generation}
offers a detailed description of the logical
hierarchy graph generation process used
in GCN-based model training. 
Section~\ref{sec:experimental_setup} discusses
the experimental setup, including data
generation, data separation for model
training and testing, and the details of
the model training process.
Section~\ref{sec:experimental_results}
presents our experimental results, which
include an assessment of different sampling
methods, the performance evaluation of
our prediction models, and the results from
DSE. Finally, Section~\ref{sec:conclusion}
concludes the paper and outlines future
research directions.
}

\section{Related work}
\label{sec:related_work}
\noindent
Prior efforts have sought to predict power, performance,
and area (PPA) at different stages of the design flow using two
classes of predictors, respectively based on analytical
models and ML models.
The works in~\cite{KahngLN15, LeeKRSGJ15} introduce ML models and
demonstrate significant improvement in PPA prediction
over previous analytical models such as
ORION~\cite{WangZPM02} and McPAT~\cite{LiASBTJ09}. 
Works include power metric
prediction for high-level synthesis (HLS)~\cite{DaiZZUYZ18}
and PPA prediction for memory compilers~\cite{LastS21}.
Such approaches indicate that ML models outperform analytical
models and are more convenient to train a wide range
of ready-to-use ML models. To the best of our knowledge, no prior
work builds ML models based on full backend SP\&R for ASIC.
In~\cite{BaiSZMYW21}, an NN-based ML model is used to predict power
and performance for different microarchitectures and
to find
Pareto optimal design points for power and performance.
An ML model to estimate power
metrics in HLS, along with sampling-based techniques to prune the
search space, is presented in~\cite{LinZSZ20};
these methods are used to find the Pareto frontier and
Pareto-optimal designs for FPGA implementations.

\sk{Recently, several studies have showcased the use of GNNs 
for predicting PPA, although they do not focus on ML hardware. 
The work
in \cite{SenguptaTCH22} directly predicts post-placement power and
performance from RTL, demonstrating that the XGBoost model surpasses the
performance of GCN-based models. 
The authors of~\cite{SenguptaTCH22} report 95\% accuracy in terms 
of R-squared score, but calculation of the absolute percentage 
error from the provided data reveals a significant prediction error.
Several other studies have employed GNNs for PPA prediction.
For instance, \cite{LuCKL22} forecasts
final PPA from the early stages of the P\&R flow;
\cite{LiWLLL22}
predicts post-synthesis PPA for Network-on-Chip (NoC) design using NoC
parameters, topology configurations, and task graphs with the aid of a
message-passing neural network; and \cite{cheng23} employs the graph
representation of the netlist to predict leakage recovery during the 
ECO stage. Nevertheless, these studies do not offer prediction for
post-route optimization or backend PPA from the RTL. In contrast,
our work takes a novel approach for a given ML accelerator. We 
generate a logical hierarchy graph from the RTL and utilize GNNs to 
extract graph embeddings, which are then employed for backend
PPA prediction.}

In the ML hardware context, there is limited prior work on
the early prediction of DNN
performance. Aladdin~\cite{ShaoRWB14} combines PPA-characterized
building blocks with a dataflow graph representation to estimate
performance, but does not incorporate the impact of physical
design (PD) decisions beyond the block level. However, these decisions
can substantially impact system performance.
NeuPart~\cite{Susmita21} develops an analytical model to predict 
energy for computation and communication in a DNN accelerator.
AutoDNNchip~\cite{XuZHZZW20} proposes a predictor for energy,
throughput, latency, and area overhead of DNN accelerators based
on architectural parameters. It determines system-level
performance metrics in an analytical-model-based coarse-grained
mode and a runtime-simulation-based fine-grained mode, but has
no clear engagement with backend design optimizations.
On the other hand, it is well-understood that the performance
of an ML accelerator is acutely dependent on the tradeoffs made
in backend design. Numerous technology, methodology, and
tool/flow effects must be comprehended, modeled and exploited --
e.g.,~\cite{AgnesinaCL20} shows that post-routing wirelength
can be improved by about 15\% with different settings of
flow knobs. ML accelerators are considerably more complex
than the testcases used in \cite{AgnesinaCL20}, and can be expected to show
even greater overall variation in post-routing outcomes due to
tool/flow effects. In contrast to all previous works, our approach
takes the effect of backend flows into account and effectively
ties the prediction of {\em system-level} performance metrics to backend
PPA.

\clearpage
\section{Key definitions and notation}
\label{sec:notations}
We formally define a set of key terms used in our work.
\begin{itemize}[noitemsep,topsep=0pt,leftmargin=*]
   \item {\bf ML accelerator (design).} An ML accelerator
   or design is the RTL netlist
   created by a parameterizable ML hardware generator.
   \item {\bf Workload.} A workload is a user-specified ML 
   algorithm (or a set of algorithms) that runs on an ML 
   accelerator. Due to the inherent structure of
   the computation for a given network, the cost metrics
   for a workload -- i.e., the energy and runtime of the
   accelerator -- depend on the network topology and not
   on the specific input data.
   \item {\bf Architectural parameters.} These are a set of parameters 
   used by a parameterizable ML hardware generator to generate an ML accelerator.
   A \textbf{configuration} is a specific setting of architectural parameters.
   The parameterizable ML hardware generator can only generate
   one ML accelerator for a given configuration.
   This means there is a one-to-one mapping between ML 
   accelerators and configurations for a 
   parameterizable ML hardware generator.
   \item {\bf Target clock period.} The target clock period 
   is the clock period in the .sdc
   (Synopsys Design Constraints) file.
   The target clock frequency
   ($\tcf$) is the reciprocal of target 
   clock period. \sk{Altering $\tcf$ impacts the outcome of
   the SP\&R process. SP\&R tools aim to meet the specified
   $\tcf$, neither more nor less. Exceeding the $\tcf$ can 
   result in increased power consumption and area usage,
   whereas falling short of the $\tcf$ will lead to
   compromised performance.}
   \item \sk{{\bf Floorplan utilization.} Floorplan utilization ($util$) is 
   an input parameter within the backend flow that determines the chip 
   area based on the input synthesized netlist. The chip area is typically
   calculated as total standard cell and macro area within the
   synthesized netlist, divided by the floorplan utilization.}
\end{itemize}
For an ML accelerator:
\begin{itemize}[noitemsep,topsep=0pt,leftmargin=*]
   \item {\bf Power ($P$)} is the sum of internal,
    switching, and leakage power, as reported by the
    SP\&R tool after post-routing optimization.
   \item {\bf Performance} is the effective clock frequency ($\ecf$).
    This is the reciprocal of effective clock period, 
    defined as the target clock period minus the worst slack reported 
    by the SP\&R tool after post-routing optimization.
    \item \sk{{\bf Area ($A$)} is the chip area reported by the P\&R 
    tool. 
    For all our runs, we consider a chip aspect ratio of one,
    implying that the chips are square in shape.}
    \item {\bf Energy ($E$)} is the total energy required to run 
    the user-specified workload on the ML accelerator.
    Given an ML accelerator and a workload,
    a simulator computes the energy 
    based on the instruction mix and the post-SP\&R 
    performance/power metrics of the 
    accelerator submodules.
    \item {\bf Runtime ($T$)} is the time required to 
    run the user-specified workload.
    For an ML accelerator and a specific workload,
    a performance simulator is used to compute the runtime.
\end{itemize}

\section{Problem Formulation}
\label{sec:problem_formulation}
\subsection{Prediction framework}
We divide the problem of full-stack optimization of ML
hardware accelerators into two subproblems.
Given an ML architectural 
configuration $({x_1, x_2, \cdots, x_n})$, 
\sk{the logical hierarchy graph ($LHG$)} of the RTL netlist,
the target clock frequency ($\tcf$) and the
floorplan utilization ($util$), two subproblems are as follows.
\sk{
\begin{itemize}[noitemsep]
    \item \textbf{Problem 1:} predict backend power ($P$), performance ($f$) and area ($A$). 
    \item \textbf{Problem 2:} predict system-level runtime ($T$) and energy ($E$).
\end{itemize}
}

\noindent
To solve above two problems, we build two
learning-based prediction frameworks:
\begin{itemize}
    \item The first predicts backend power ($P$), performance ($f$) and area ($A$) 
    \begin{align}
        (P, f, A) = ML Model (\{x_1, x_2, \cdots , x_n; \sk{LHG}, \tcf, \sk{util}\})
    \label{eq:ML4MLbackend}
    \end{align}
    \item The second predicts system-level runtime ($T$) and energy ($E$)
    \begin{align}
        (E, T) = ML Model (\{x_1, x_2, \cdots , x_n; \sk{LHG}, \tcf, \sk{util}\})
    \label{eq:ML4MLsystem}
    \end{align}
\end{itemize}

\noindent
\sk{We propose to leverage the RTL netlist produced by the ML
hardware generator, together with architectural and
backend features, as a strategy to enhance performance
of our ML model. We apply the Graph Convolutional
Network (GCN) modeling, which is adept at extracting
relevant features from a graph. We generate an $LHG$ from the
RTL netlist, and incorporate this as an additional input into our
GCN model, alongside the architectural and backend features.}

\subsection{Design Space Exploration}
\sk{Next, we define the problem of design space exploration.
Specifically, for a given target workload, our objective is to 
figure out the optimal architectural and backend configuration,
which minimizes the cost function:
\begin{equation}
\label{eq:DSE}
    (\alpha\times E + \beta\times A)
\end{equation}
while satisfying following constraints.
\begin{itemize}[noitemsep]
    \item The total power ($P$) is less than the specified maximum power $P_{max}$ ($P < P_{max}$).
    \item The runtime ($R$) is less than the specified maximum runtime $R_{max}$ ($R < R_{max}$).
    \item The energy ($E$) and area ($A$) belongs to the Pareto front
    of ($E$, $A$) ($(E, A)\in \textit{Pareto Front}(E, A)$).
\end{itemize}

\noindent
Here, $\alpha$ depends on the expected lifespan of the chip,
and $\beta$ depends on the fabrication cost of the chip.}

\section{Overview of our approach}
\label{sec:approach}
\begin{figure}[!htb]
    \centering
    \includegraphics[width=1.0\columnwidth]{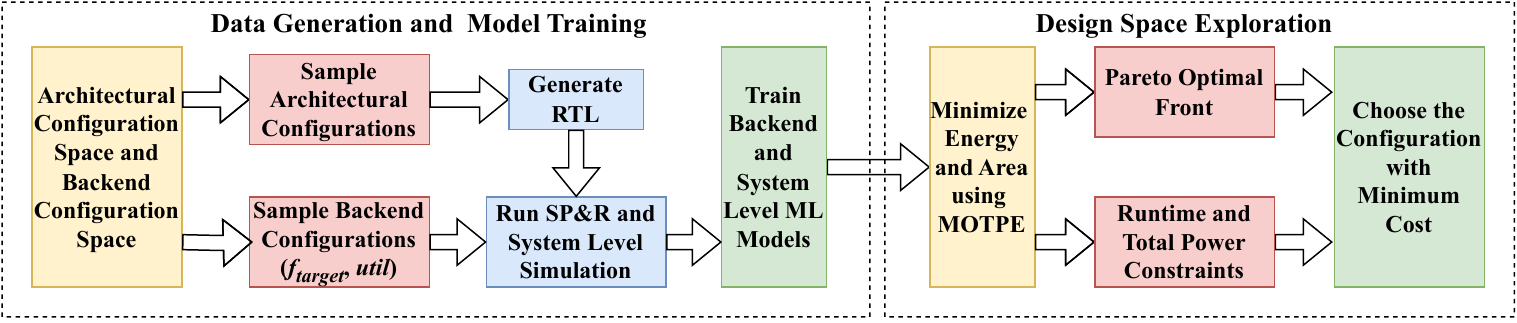}
    \caption{\sk{The ML-based full-stack optimization framework for machine learning accelerators.}}
    \label{fig:flow}
\end{figure} 
\sk{We first sample architectural configurations
and backend configurations using
different sampling methods, e.g.,
Latin Hypercube sampling, 
and Halton and Sobol 
sequence-based sampling. We generate RTL netlist
for each sampled architectural configuration
and run. Each RTL netlist goes through the SP\&R and
system-level simulation flow for each sampled
backend configuration and backend PPA metrics,
system-level metrics are captured for
model training and testing. We use this
dataset to train different ML models, e.g.,
XGBoost (XGB), Random Forest (RF), Artificial
Neural Network (ANN) and Graph Convolutional
Network (GCN).}

\sk{For a given set of architectural and backend configuration ranges,
we use trained models and the Multiobjective Tree-structured Parzen
Estimator (MOTPE) to optimize energy and chip area, adhering to power
and runtime constraints. We identify the Pareto optimal front for
energy and chip area metrics, and select the best configuration according to
Equation~\eqref{eq:DSE}.}

\subsection{Demonstration Platforms and Simulators}
\label{subsec:platform-simulators}
We demonstrate our approach on the accelerator engines from four
parameterizable open-source ML hardware generators.

\smallskip
\noindent
{\bf TABLA}~\cite{TABLA} implements non-DNN ML algorithms such as linear
regression, logistic regression, support vector machines (SVMs), backpropagation,
and recommender systems. 

\smallskip
\noindent
{\bf GeneSys}~\cite{RTML} executes DNNs using
an $M \times N$ systolic array for GEMM operations
such as convolution, and an $N \times 1$ SIMD array for vector operations such
as ReLU, pooling, and softmax.  

\smallskip
\noindent
{\bf Axiline}~\cite{RTML, ZengS23} builds hard-coded implementations of
small ML algorithms (e.g., SVM, logistic/ linear regression) for
training and inference.

\smallskip
\noindent
{\bf VTA}~\cite{VTA, IntelVTA} is a DNN accelerator where 
a compute module includes a GEMM core for convolution,
along
with a 2D array of PEs. VTA is integrated with Apache 
TVM~\cite{TVM2018}, a deep learning compiler stack.

\smallskip
\noindent
{\bf Integrating system simulations with backend data.}
The simulators are integrated with the backend analyses,
where they receive
PPA characteristics generated by our SP\&R flow.
The simulators used in our study are obtained from the GitHub
repository of the VeriGOOD-ML project~\cite{VeriGOOD-ML}
and VTA hardware design stack~\cite{IntelVTA-repo}.
For a specific hardware configuration point, provided as an input, the PPA
characteristics feed the simulator with data such as the clock frequency,
energy per access for each of the on-chip buffers,
and dynamic and leakage power of systolic and SIMD hardware components,
or GEMM and ALU hardware components.
The performance statistics provided by the simulator are combined
with these backend data from the SP\&R flow to
produce end-to-end runtime, energy, and power
for execution of a given ML algorithm.

\subsection{Sampling Method}
\sk{Sampling is a very important step for data generation.
Improper sampling leads to an unbalanced
dataset, resulting in poor performance of the trained models.
Moreover, sampling techniques can 
help reduce the number of data points needed to train
a model without degrading the performance. In this work,
we have studied three sampling techniques, including
(i)~Latin Hypercube sampling (LHS), (ii)~low-discrepancy
sequence (LDS) using Sobol and (iii)~LDS using 
Halton.}

\noindent
\sk{{\bf Latin Hypercube sampling.} LHS divides the parameter
space into equally spaced intervals along each dimension
and then randomly selects points in the intervals such that
each interval is selected exactly once. 
During the sampling process we 
maximizes the minimum pairwise
distance of the sampled points.}

\noindent
\sk{{\bf LDS using Halton.}
An LDS, also known as
a quasi-random sequence, is a sequence of points in a
multi-dimensional space that exhibits more uniform and
evenly distributed patterns, as compared to random
sequences. These sequences are generated using
deterministic algorithms that carefully distribute
points across the space to minimize undesirable
patterns. The Halton~\cite{lds} sequence relies on
unique prime number bases to generate uniformly
distributed samples.}

\noindent
\sk{{\bf LDS using Sobol.}
The Sobol~\cite{lds} sequence is also an LDS which
utilizes primitive polynomials and bitwise operations 
to generated uniformly distributed samples. 
In the case of a high-dimensional
parameter space, the Sobol sequence is expected
to exhibit a more uniform distribution 
with a smaller sample size than the
Halton sequence.}

\sk{
We use the scikit-optimize package~\cite{scikitOptimize} 
to generate samples for all three sampling methods.
Both LHS and LDS yield superior results compared to
random sampling, particularly when dealing with
smaller sample sizes. One of the benefits of LHS
is its ability to uniformly sample from the
parameter space with a smaller sample size than
LDS. However, a limitation of LHS is that
to increase the number of samples,
one needs to regenerate all the samples
and cannot reuse the previously sampled data
points. This is because adding new points
to existing LHS samples disrupts the LHS
property of maximizing the minimum pairwise
distance between sampled points.
On the other hand,
LDS, which generates samples from a sequence,
only needs to extend the sequence to generate
new samples. This feature allows LDS to utilize
previously sampled data points when the sample
size is increased.
}

\subsection{ML Models}
\sk{In our framework, we employ a range of regression models.
A brief description of each model follows.}
\begin{itemize}[noitemsep,topsep=0pt,leftmargin=*]
    \item {\bf Gradient Boosted Decision Trees (GBDT)}
    utilize multiple decision trees as weak predictors.
    New trees are added sequentially to minimize the
    loss function during the training process.
    \item {\bf Random Forest (RF)} also uses decision
    trees, but trains each tree independently using
    random samples of data. The final decision
    is generated based on voting over a set of
    trees or by averaging the predictions
    generated by each tree.
    \item {\bf Artificial Neural Network (ANN)} is a
    biologically-inspired model consisting
    of multiple neuron or node layers: an input layer,
    one or more hidden layers, and an output layer
    consisting of a single node. The output of
    any node is a linear transformation of the previous
    layer input followed by a non-linear or linear activation
    function.
    \item{\bf Stacked Ensemble} 
    uses multiple ``base
    learner'' algorithms to outperform each of the individual
    base learners. Training entails
    (i) training of multiple base learners (e.g., RF and GBDT models); and
    (ii) training of a second-level ``meta learner'' to find the optimal
    combination of the base learners as the stacked ensemble model. Theorem 1 in 
    \cite{LaanPH07} proves that the stacked ensemble model will asymptotically perform as well as the best learner. 
    \item \sk{{\bf Graph Convolutional Network (GCN)} 
    performs convolution
    operations on graphs to capture
    structural dependencies and relationships inherent within the data. This is
    accomplished by iteratively propagating and aggregating information
    throughout the graph, enabling the model to learn from and adapt to
    the complex interconnected structure of the input data.}
\end{itemize}

\sk{
We train GBDT, RF and ANN models using the open-source 
platform H2O~\cite{H2O}.
H2O enables fast, distributed, 
in-memory, and scalable machine learning-based model 
training. 
In Section~\ref{subsec:model-training},
we provide a detailed 
step-by-step guide on how to train GBDT, RF, ANN and
Stacked Ensemble models.
We train GCN using PyTorch~\cite{pyTorch} and PyTorch Geometric~\cite{pyG}.
In Section~\ref{sec:graph_generation},
we provide a detailed step-by-step guide on how to generate
graphs, and in Section~\ref{subsec:model-training}, we outline 
the process for training the models.}

\begin{figure}[!htb]
    \centering
    \begin{subfigure}[]
    {\includegraphics[width = 0.51\columnwidth]{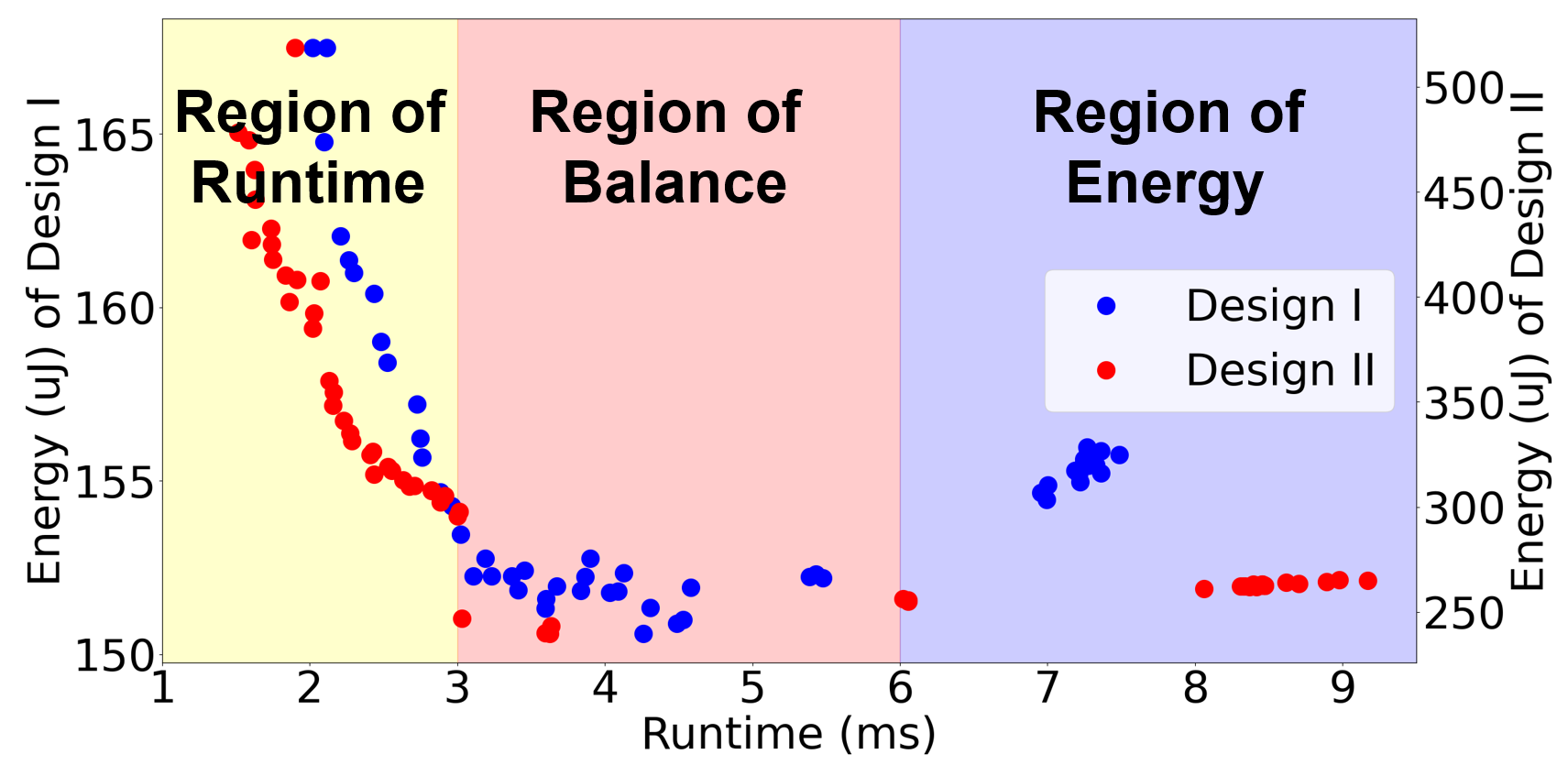}}
    \end{subfigure}%
     \begin{subfigure}[]
    {\includegraphics[width = 0.48\columnwidth]{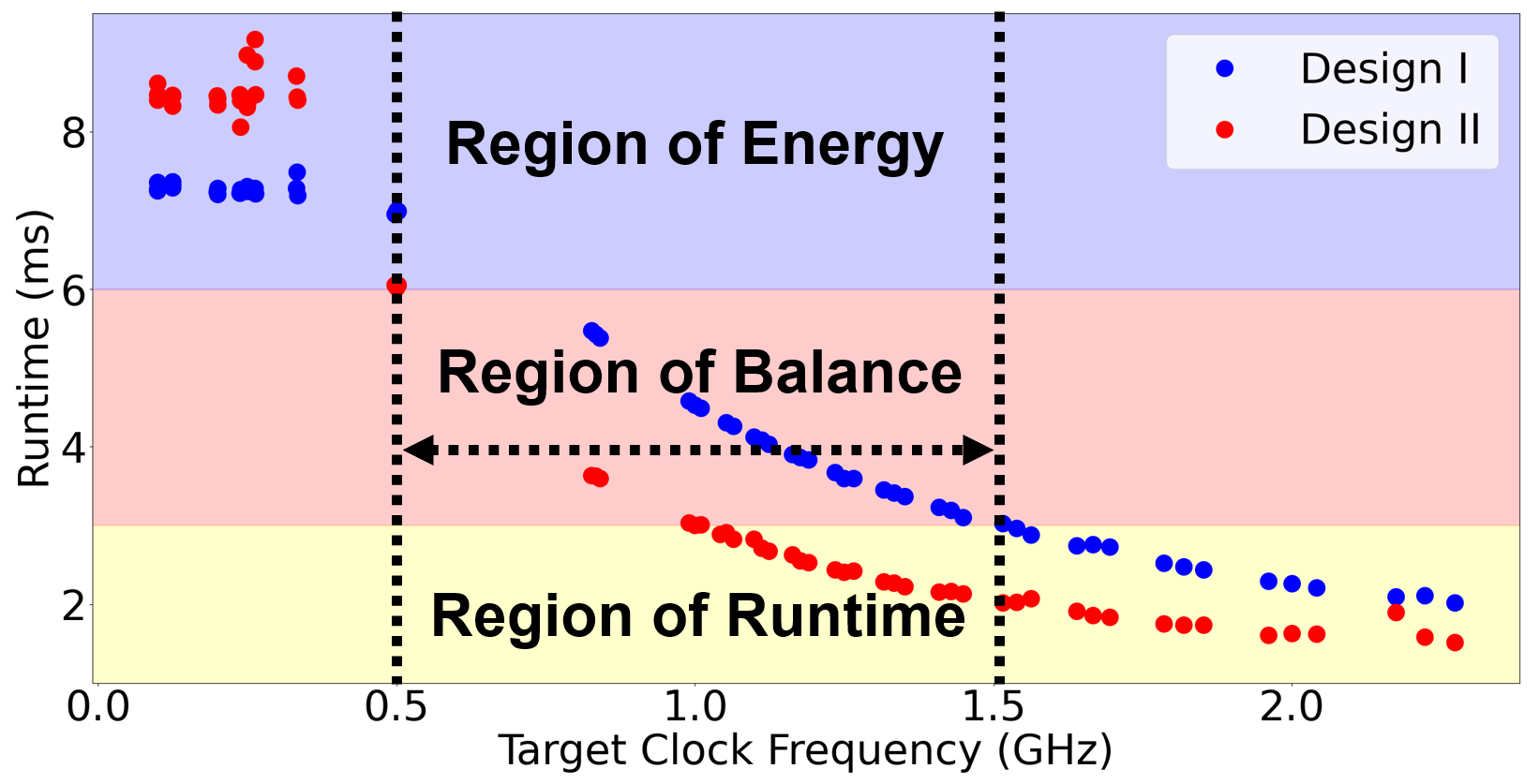}}
    \end{subfigure}%
    \begin{subfigure}[]
    {\includegraphics[width = 0.49\columnwidth]{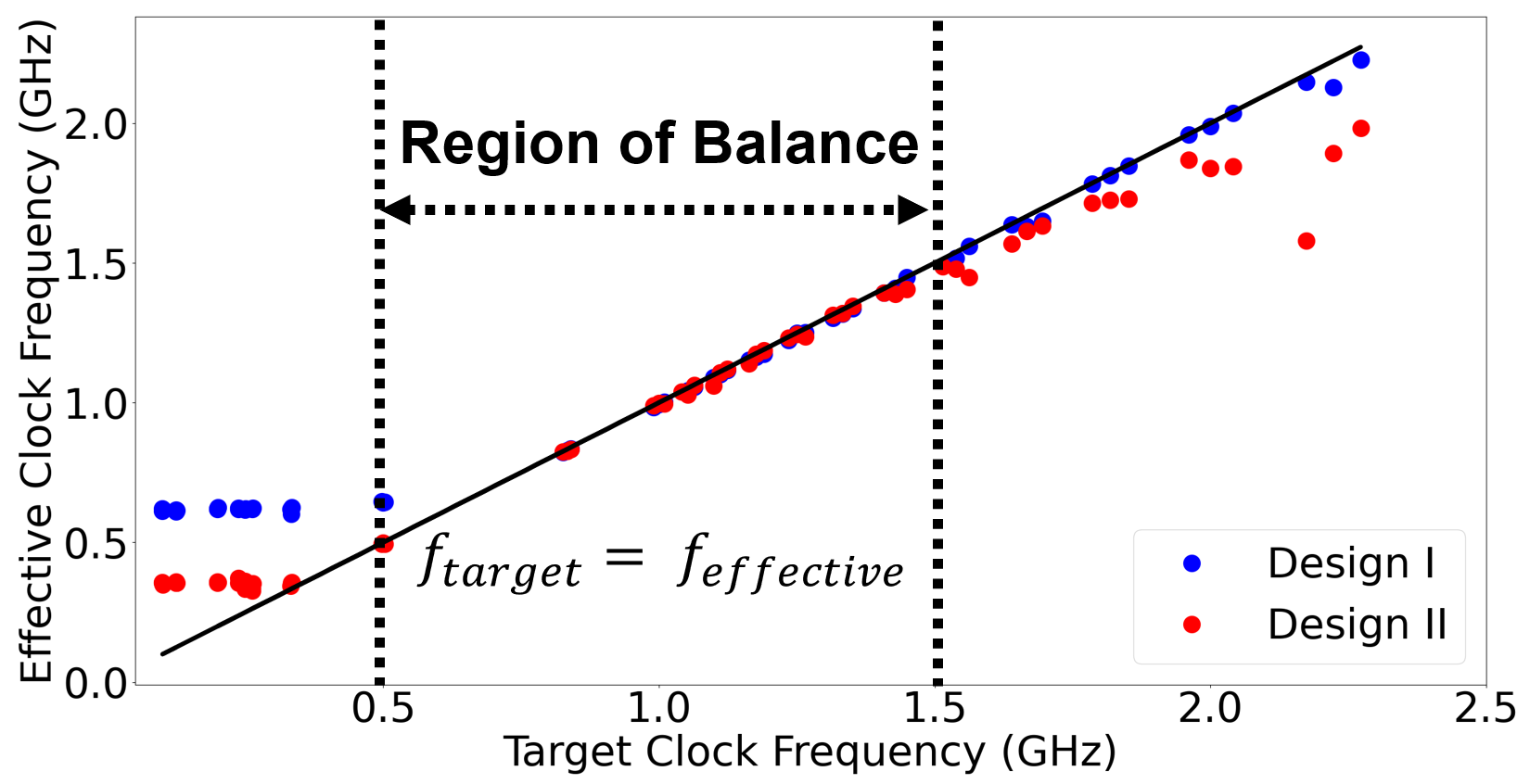}}
    \end{subfigure}
    \caption{Illustration of the region of interest (ROI).
    (a) Energy versus runtime. The ROI is the pink region.
    (b) Runtime versus target clock frequency. Again, the ROI is the pink region.
    (c) Target clock frequency versus effective clock frequency. The ROI is the region where 
    $\tcf = \ecf$.}
    \label{fig:power_error}
\end{figure}

\subsection{Two-Stage Model}
\label{subsec:two-stage}
Design space exploration seeks an optimal implementation
of an accelerator for a specified workload, 
to achieve better backend PPA and system metrics.
To explore the design space more efficiently,
we pay more attention to a {\em region of interest} (ROI) in the design space,
i.e., the region that contains the ``optimal'' implementation.
Figure~\ref{fig:power_error} gives an example of how to determine the ROI.
In the example, we first use the Axiline platform to generate two accelerators 
({\em Design-I} and {\em Design-II}) with different configurations, to implement a recommender system algorithm.
We then run full SP\&R flows for the two accelerators under 21 different 
$\tcf$ values, and compute corresponding energy and runtime system metrics.
Figure~\ref{fig:power_error}(a) shows energy versus runtime
of {\em Design-I} and {\em Design-II} for different $\tcf$.
The data show a typical division of the design space into three regions:
(i) region of {\em runtime} where we can reduce the
runtime at the cost of increasing energy;
(ii) region of {\em balance} where we can achieve lowest energy without 
introducing too much runtime overhead; and
(iii) region of {\em energy} where the energy will also
increase when the runtime increases.
In our work, we set the ROI to be the region of balance.
To visualize the ROI in terms of $\tcf$, we show 
plots of runtime versus $\tcf$ in Figure~\ref{fig:power_error}(b).
We observe that the ROI excludes both extremely high and low $\tcf$.
Then, by examining $\ecf$ versus $\tcf$ 
(Figure~\ref{fig:power_error}(c)), we may further characterize the ROI 
in terms of the difference between $\tcf$ and 
$\ecf$:  post-routed designs tend to have smaller negative 
slacks 
at higher $\tcf$ 
and larger positive slacks at lower $\tcf$.
Given the above, we define our ROI in terms of the 
difference between $\tcf$ and $\ecf$,
as follows:
\begin{equation}
\label{eq:ROI}
    ROI = \{ \tcf  | abs(\ecf - \tcf) \leq \epsilon \times \tcf \}
\end{equation}

\begin{figure}[!htb]
\centering
\begin{subfigure}[]
{\includegraphics[width = 0.49\columnwidth]{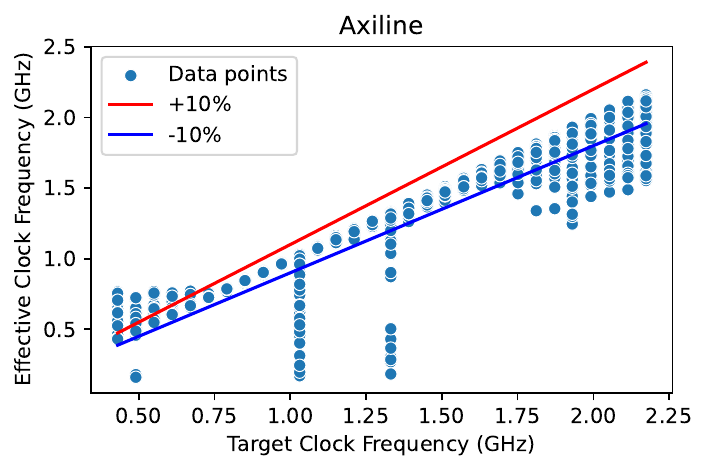}}
\end{subfigure}
\begin{subfigure}[]
    {\includegraphics[width = 0.49\columnwidth]{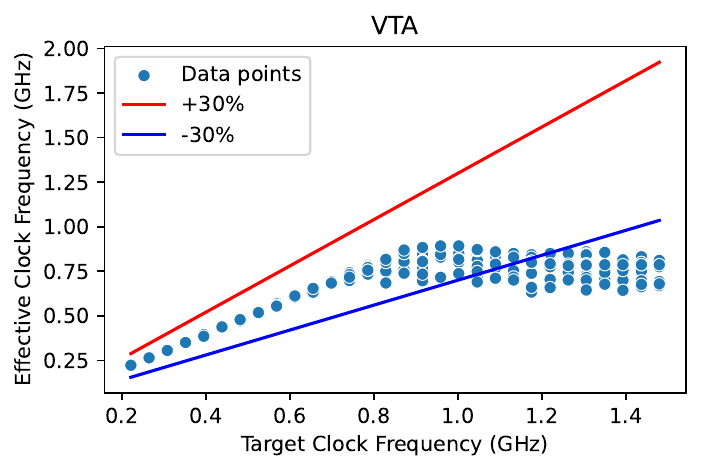}}
\end{subfigure}
\begin{subfigure}[]
    {\includegraphics[width = 0.49\columnwidth]{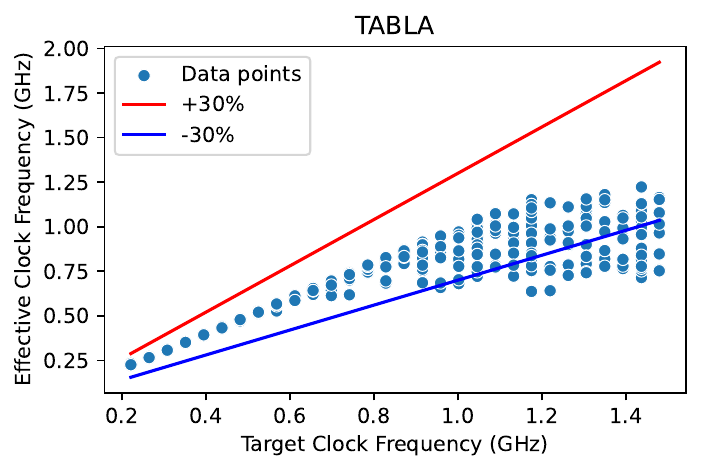}}
\end{subfigure}
\caption{\sk{Effective clock frequency vs. target clock frequency plot 
for (a) Axiline, (b) VTA and (c) TABLA designs on GF12 enablement.
In this case, the floorplan utilization is not the same for all $\tcf$ values;
it varies as shown in Figure~\ref{fig:backend_configurations}.}}
\label{fig:axiline_ecp_vs_tcp}
\end{figure}

\sk{
We include floorplan utilization as a separate knob to control
chip area. We observe that high utilization significantly
impairs the outcomes of the backend P\&R tool, leading to
poor PPA. Figure~\ref{fig:axiline_ecp_vs_tcp}(a) presents
the $\ecf$ vs. $\tcf$ plot for the Axiline design on GF12.
For $\tcf$ values of 1.03 and 1.33 GHz, the
floorplan utilization hovers around 90\%, which results in
poor postRouteOpt performance for most of the Axiline designs.
Additionally, when the $\tcf$ is very high, the P\&R tool
struggles to achieve the desired performance, resulting in
a poor $\ecf$. Conversely, when $\tcf$ is very low, the tool
yields a higher $\ecf$ than $\tcf$. These extreme $\tcf$
values are also undesirable as the outcomes from the
P\&R tool for these frequencies tend to vary significantly,
making them challenging to model accurately. So, these data
points are considered as outliers, since they
do not correspond to relevant design points, and since
their inclusion in the training
dataset deteriorates the performance of the model.}
  
\sk{We modify the two-stage inference model proposed in
\cite{Esmaeilzadeh22} into a two-stage model based on the
ROI definition above. The goal is to efficiently
detect outliers and prevent them from impacting the
performance of the model. The two-stage model functions as
follows: (i)~we first train a binary classification model to
detect the data points that belong to the ROI; (ii)~we then
train all our models to predict PPA only for the data points
that belong to the ROI; (iii) we employ these trained models
to predict PPA solely for the predicted data points that belong
to the ROI according to the classification model; and (iv) if the
classification model predicts that a data point does not belong
to the ROI, we discard this data point. In Equation~\eqref{eq:ROI},
$\epsilon$ is a parameter used to define the size of the ROI.
For smaller ML accelerators such as Axiline, where the deviation of
the $\ecf$ from the $\tcf$ is generally small, we set $\epsilon$
to 0.1. For larger ML accelerators such as GeneSys, VTA and
TABLA, where the deviation of $\ecf$ from $\tcf$ tends to be
larger, we set $\epsilon$ to 0.3.
}

\subsection{Design Space Exploration}
\sk{In our methodology, we utilize the above two-stage
model for DSE of both architectural and backend configurations. During
this DSE process, our initial step involves the identification of
the Pareto front of backend metrics (power, performance and area) and 
system-level metrics (runtime and energy). This is accomplished
via the multi-objective tree-structured Parzen estimator
(MOTPE). Subsequently, the optimal configuration is selected
according to a user-specified cost function.}

\noindent
\sk{{\bf MOTPE-based DSE.}
MOTPE~\cite{motpe} is a sequential model-based optimization technique.
It iteratively builds a surrogate model to predict performance
and gather additional data informed by this model.
The estimator constructs distributions of good ($G$) and
bad ($B$) samples, categorizing samples based on their 
relative position in the objective space with respect to
the current Pareto front. Before forming these internal
distributions, the algorithm collects multiple random 
samples. It utilizes a non-parametric multivariate density
estimation model, commonly referred to as the Parzen window,
to construct these distributions. Subsequently, the MOTPE
selects the sample which maximizes the $G/B$ ratio,
which ensures superior sampling efficiency. A significant 
advantage of MOTPE is its capacity to handle both discrete
and continuous-valued parameters, thereby facilitating its
application across a broad range of optimization scenarios.
When conducting DSE of ML accelerators, backend parameters
such as floorplan utilization and target clock frequency are
continuous in nature, whereas architectural parameters such
as the number of processing units and processing elements are
discrete. This makes
MOTPE-based optimization well-suited to find the 
Pareto front points of PPA and system-level metrics of energy
and runtime. Based on the desired configuration and cost function,
we choose the final architectural and backend configuration
of the ML accelerator. 
Identifying the Pareto front allows us to understand the 
tradeoffs among different metrics, which is independent of
user-specified weights in the cost function (i.e., $\alpha$ and $\beta$
in Equation~\eqref{eq:DSE}).}

\section{Graph Generation}
\label{sec:graph_generation}

\sk{ML accelerators exhibit a high degree of modularity in 
their design. For different architectural configurations, the 
ML hardware generators leverage various building blocks to 
generate the RTL netlist. We extract information about these 
building blocks and their connections from the RTL netlist. A graph 
representation of the connection between these building blocks 
is utilized in training our GCN models, which helps enhance 
model performance. For the GCN model, we extract the logical 
hierarchy graph (LHG) from the RTL netlist. 
The specifics of the logical hierarchy graph are as follows.
}
  
\begin{figure}[htbp]
    \centering
    \includegraphics[width=0.6\columnwidth]{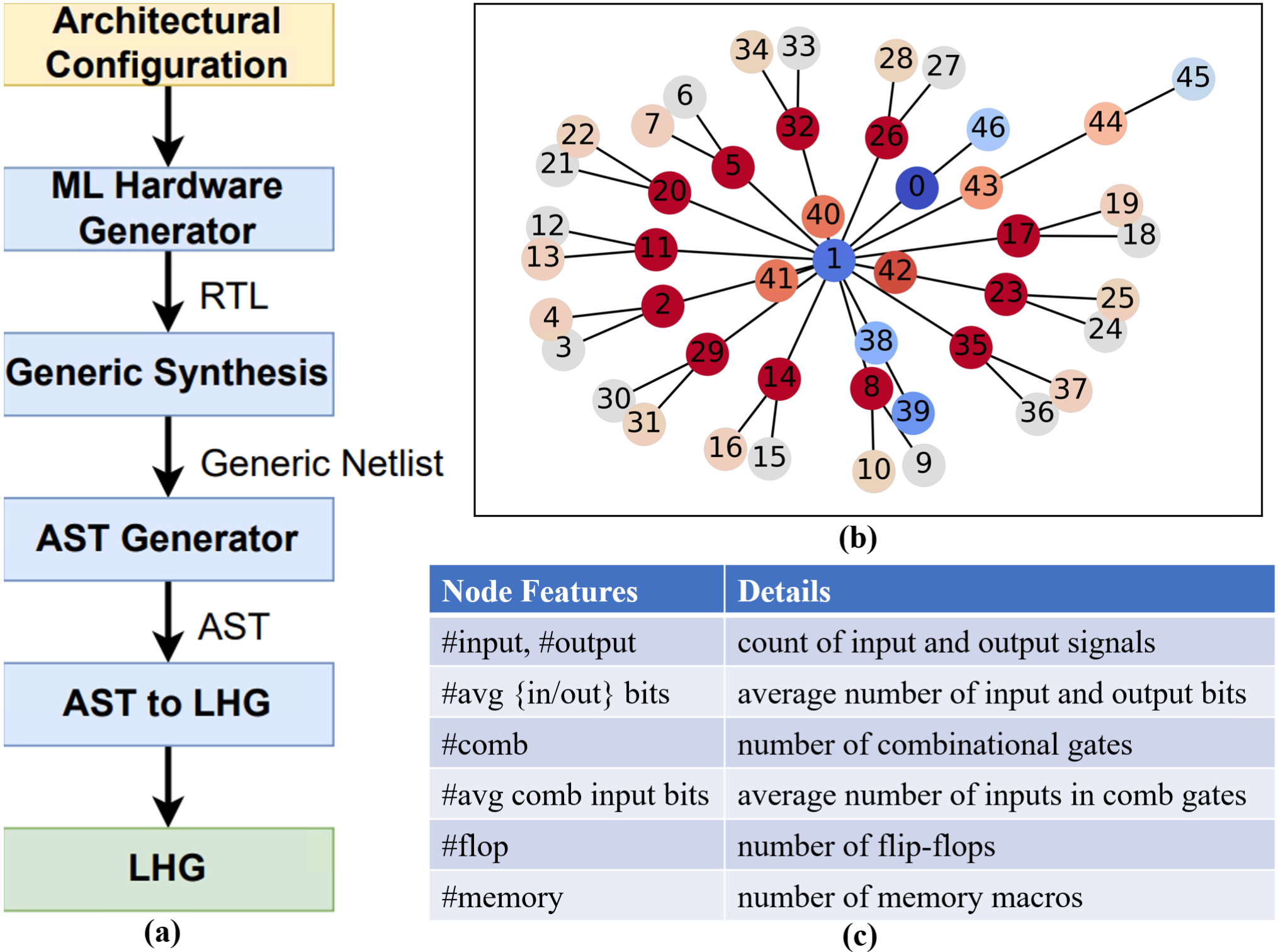}
    \caption{\sk{(a) Logical hierarchy graph generation flow. 
    (b) Example of logical hierarchy graph of an Axiline design.
    Here, nodes with the same color correspond to the building 
    blocks with the same
    functionality. (c) Details of node features.}}
    \label{fig:lhg_flow}
\end{figure}

\noindent
\sk{{\bf Logical hierarchy graph (LHG).} 
As the name suggests, the LHG represents the 
logical hierarchy tree of the design for a 
given architectural configuration. 
Each module instantiation in the RTL netlist
is mapped to a node in the LHG. Each undirected 
edge connects a parent module to its sub-module. All the leaf modules are 
the building blocks of the ML hardware generator. Figure~\ref{fig:lhg_flow}(a) 
depicts the flow of generating an LHG for a given
architectural configuration.}

\sk{
\begin{itemize}[noitemsep,topsep=0pt,leftmargin=*]
    \item We first generate the RTL netlist for a given
    architectural configuration, using an RTL generator
    such as VTA, GeneSys, TABLA or Axiline.
    \item Next, we transform the RTL netlist into a
    structural netlist, referred to as a {\em generic
    netlist}, using Cadence Genus 21.1.
    \item Using Pyverilog~\cite{pyverilog, iverilog}, we
    generate the abstract syntax tree (AST) of the generic
    netlist.
    \item From the AST, we extract node features and generate the LHG using Algorithm~\ref{algo:lhg}.
\end{itemize}
}

\begin{algorithm}[htbp]
\caption{\sk{Generate Logical Hierarchy Graph from AST.}}
\label{algo:lhg}
\KwInput{AST, TopModuleName \\}
\KwOutput{Logical hierarchy graph $G$ \\}
\BlankLine
Parse AST using Pyverilog and create a list of unique modules \\
Extract features for each module and create a reference node list \\ 
Update the list of child nodes in the reference node list \\
$G \gets$ Empty Graph; $id \gets 0$; $pid \gets -1$ \\
{\em AddNodeToGraph}(\textit{refNode} of TopModule, $G$, $pid$, $id$) \\
\textbf{return} $G$ \\

\nonl\textbf{Procedure:} {\em AddNodeToGraph} \\
\KwInput{refNode, $G$, $pid$, $id$ \\}
\KwOutput{$id$ -- is the node count of the updated graph $G$ \\}
\setcounter{AlgoLine}{0}
\BlankLine
$G.addNode(id, refNode)$ \\
$node\_id \gets id$; $ id \gets id + 1$ \\
\If{$pid \neq -1$} {
    $G.addEdge(pid, node\_id)$
    }

\For{$subModuleNode$ {\bf in} $refNode.subModuleList$} {
$id \gets AddNodeToGraph(subModuleNode, G, node\_id, id)$
}
{\bf return} $id$
\end{algorithm}

\sk{
The generic netlist serves as a structural representation of the RTL netlist,
composed of Verilog primitives like $or$, $and$ and $inv$ for 
combinational logic, as well as dedicated modules such as flip-flops
and latches for sequential logic. We generate and parse the AST of 
the generic netlist using Pyverilog~\cite{pyverilog, iverilog}.
Algorithm~\ref{algo:lhg} displays
the detailed steps of LHG generation. Initially, we list all the modules
of the RTL netlist and extract the 
node features illustrated in Figure~\ref{fig:lhg_flow}(c).
These features include input and
output signal counts, average number of input and output bits, combinational 
cell count, flip-flop count, memory count and the average number of inputs 
for combinational cells. Importantly, these features rely solely on the
RTL netlist and not on the backend parameters, i.e., $\tcf$ and $util$;
therefore, changing the backend configuration does not require updating
the LHG.}

\sk{The advantage of using a generic netlist instead of the RTL netlist is that
we can directly extract the number of combinational cells and the 
flip-flop count for each module, which is not possible
with the AST
generated directly from the RTL.
During the synthesis process, the RTL
netlist is mapped to a specific design library
based on the given {\em Synopsys Design Constraints}.
In contrast, during the generic netlist generation 
process, the RTL netlist is transformed to a
structural format. So, the time taken to
generate the generic netlist is significantly less than the
synthesis runtime. For example, for a GeneSys design with
900K instances, the synthesis process takes 222 minutes,
whereas the generic netlist generation process only takes 72 minutes.}

\sk{Once we have all the node features, we instantiate the graph and
utilize the {\em AddNodeToGraph} procedure to create the LHG.
The {\em AddNodeToGraph} procedure initially adds the node
corresponding to the top module, then uses depth-first search
to add the node corresponding to the submodule. When adding
the node corresponding to a submodule, it also creates an edge
between the parent module and the submodule to ensure connectivity.
Figure~\ref{fig:lhg_flow}(b) shows the logical hierarchy graph of 
an Axiline design.}
  
\sk{The number of nodes in an LHG, even for a very large ML accelerator
with millions of instances, amounts to only a few thousand.
For example, a GeneSys design with 900K instances has around
3,000 nodes in its LHG graph. The LHG graph represents the
logical hierarchy tree, so the number of edges is one less
than the number of nodes. Therefore, a high 
instance count does not pose any hindrance. 
Moreover, we opt to use graph convolutional
layers such as 
{\em GCNConv} and {\em GraphConv},
over simple GNN layers such as {\em GraphSAGE},
because graph convolutional layers possess the
capability to extract global features from the graph. Considering 
that the node count of the LHG is relatively low,
on the order of 
thousands, the use of graph convolutional layers enables
the efficient training of our
models. We can thus navigate the 
large-scale structure of ML accelerators while maintaining a 
computationally tractable model. In Section~\ref{subsec:model-training},
we include the details of the training process for the GCN model, and in 
Section~\ref{sec:experimental_results}, we discuss the performance 
of the model.
}

\noindent 

\section{Experimental setup}
\label{sec:experimental_setup}
\sk{In this section, we describe our experimental setup, which
encompasses (i) the method of data generation; (ii) the separation
of data into training and testing sets for the prediction of PPA
along with system-level metrics for unseen backend and unseen 
architectural configurations; and (iii) the detailed steps involved
in different model training procedures.}
\subsection{Data Generation}
\label{subsec:data_generation}

\sk{We divide the data generation process into three substeps: 
(i) sampling of architectural parameters and RTL netlist generation; 
(ii) sampling of backend parameters; and (iii) PPA and system-level
metric data generation.}

\noindent
{\bf Sampling of architectural parameters and RTL netlist generation.} 
All platforms, namely TABLA, GeneSys, VTA and Axiline, 
are parameterizable, with corresponding tunable architectural
parameters shown in Table~\ref{tab:features}.  
We use a variety of strategies to generate multiple configurations for each platform. 
Prior works on DNN accelerators based on 
systolic arrays~\cite{DeepOpt2020, TPU-v1} 
and vector dot-products~\cite{IntelVTA} report architectural parameters 
such as array dimension, data bitwidth, on-chip buffer size and off-chip bandwidth. 
For GeneSys and VTA, we use such insights from prior works 
to guide our choice of architectural parameters. 
We proportionally scale buffer size and bandwidth
parameters based on array dimensions.
For each array dimension, we select other architectural parameters
by changing ratios of the buffer sizes, so as to to exercise
various data reuse tradeoffs in DNNs.
For example, a larger WBUF facilitates more weight reuse 
while a larger OBUF biases a design toward more on-chip
reduction of the partial sums.  
For TABLA, we explore multiple configurations using variations
of the structures shown in~\cite{TABLA}.
For Axiline, 
we use {\em Latin Hypercube Sampling} (LHS) 
for integer architectural parameters 
such as dimension and number of cycles,
thus achieving uniform coverage in each dimension;
and we simply enumerate all the combinations 
for all other remaining architectural parameters.
\sk{In Section~\ref{subsec:asses_sampling_methods}, we conduct a
detailed analysis of the impact of various sampling methods and
sample sizes on the prediction of unseen architectural 
configurations, specifically for the Axiline design.
In Section~\ref{subsec:extrapolation}, we show
that our model underperforms when the testing dataset falls
outside the range of the training dataset. Therefore, it is
essential to ensure that the range of training dataset covers
the testing dataset.}

\begin{table}[htbp]
    \centering
    \caption{Architectural parameters for four design
    platforms.}
    \label{tab:features}
    \resizebox{\columnwidth}{!}{
    \begin{tabular}{|l|l|l|l|}
    \hline
    \textbf{Platforms}  & \textbf{Feature} & \textbf{Candidate Values} & \textbf{Description} \\  \hline
    \multirow{6}{*}{TABLA} & PU & 4, 8 & \# processing units \\ \cline{2-4}
    & PE & 8, 16 &  \# processing engines in each PU \\ \cline{2-4}
    & bitwidth & 8, 16 & bit width of internal bus \\ \cline{2-4}
    & input bitwidth & 16, 32 & bit width of {\em IO} bus \\ \cline{2-4}
    & \multirow{2}{*}{benchmark} & recommender systems & \multirow{2}{*}{ML algorithms} \\
    & &  backpropagation & \\ \hline

    \multirow{14}{*}{GeneSys} 
    & weight data width & 4 -- 8 (integer) & bit width of weight data (bit)\\ \cline{2-4}
    & activation data width & 4 -- 8 (integer)  & bit width of input activation data (bit) \\ \cline{2-4}
    & accumulation width & 32 (integer)  & bit width of output accumulation (bit) \\ \cline{2-4}
    & WBUF capacity & 16 -- 256 (integer) &  size of weight buffer (KB) \\ \cline{2-4}
    & IBUF capacity & 16 -- 128 (integer) & size of input buffer (KB) \\ \cline{2-4}
    & OBUF capacity & 128 -- 1024 (integer) & size of output buffer (KB) \\ \cline{2-4}
    & SIMD VMEM capacity & 128 -- 1024 (integer) & size of vector memory in VMEM (KB) \\ \cline{2-4}
    & WBUF AXI data width & 64 -- 256 (integer) & AXI bandwidth for the WBUF (bits/cycle) \\ \cline{2-4}
    & IBUF AXI data width & 128 -- 256 (integer) & AXI bandwidth for the IBUF (bits/cycle) \\ \cline{2-4}
    & OBUF AXI data width & 128 -- 256 (integer) & AXI bandwidth for the OBUF (bits/cycle) \\ \cline{2-4}
    & SIMD AXI data width & 128 -- 256 (integer) & AXI bandwidth for the VMEM (bits/cycle) \\ \hline
    
    \multirow{7}{*}{VTA}
	& weight data width & 8 (integer)   & bit width of weight data (bit)\\ \cline{2-4}
	& activation data width & 8 (integer)   & bit width of input activation data (bit) \\ \cline{2-4}
	& accumulation width & 32 (integer)   & bit width of output accumulation (bit) \\ \cline{2-4}
	& WBUF capacity & 16 -- 256 (integer) & size of weight buffer (KB) \\ \cline{2-4}
	& IBUF capacity & 16 -- 128 (integer) & size of input buffer (KB) \\ \cline{2-4}
	& OBUF capacity & 32 -- 512 (integer) & size of output buffer (KB) \\ \cline{2-4}
	& off-chip bandwidth  & 64 -- 512 (integer) & total external bandwidth (bits/cycle) \\ \hline
	
    \multirow{8}{*}{Axiline} & \multirow{3}{*}{benchmark} & SVM, linear regression, & \multirow{3}{*}{ML algorithms}  \\& &   logistic regression, & \\& &  recommender systems & \\\cline{2-4}
    & bitwidth & 8, 16 & bit width for computation units \\ \cline{2-4}
    & input bitwidth & 4, 8 & bit width for initial inputs \\ \cline{2-4}
    & dimension & 5 -- 60 (integer) & dimension for stages 1 and 3 \\ \cline{2-4}
    & \multirow{2}{*}{num of cycles} & \multirow{2}{*}{1 -- 25 (integer)} & number of cycles required for stages\\ & & & 1 or 3 to process one input vector\\ \hline
    \end{tabular}
    }
\end{table}

\begin{figure}
    \centering
    \begin{subfigure}[]
        {\includegraphics[width=0.485\columnwidth]{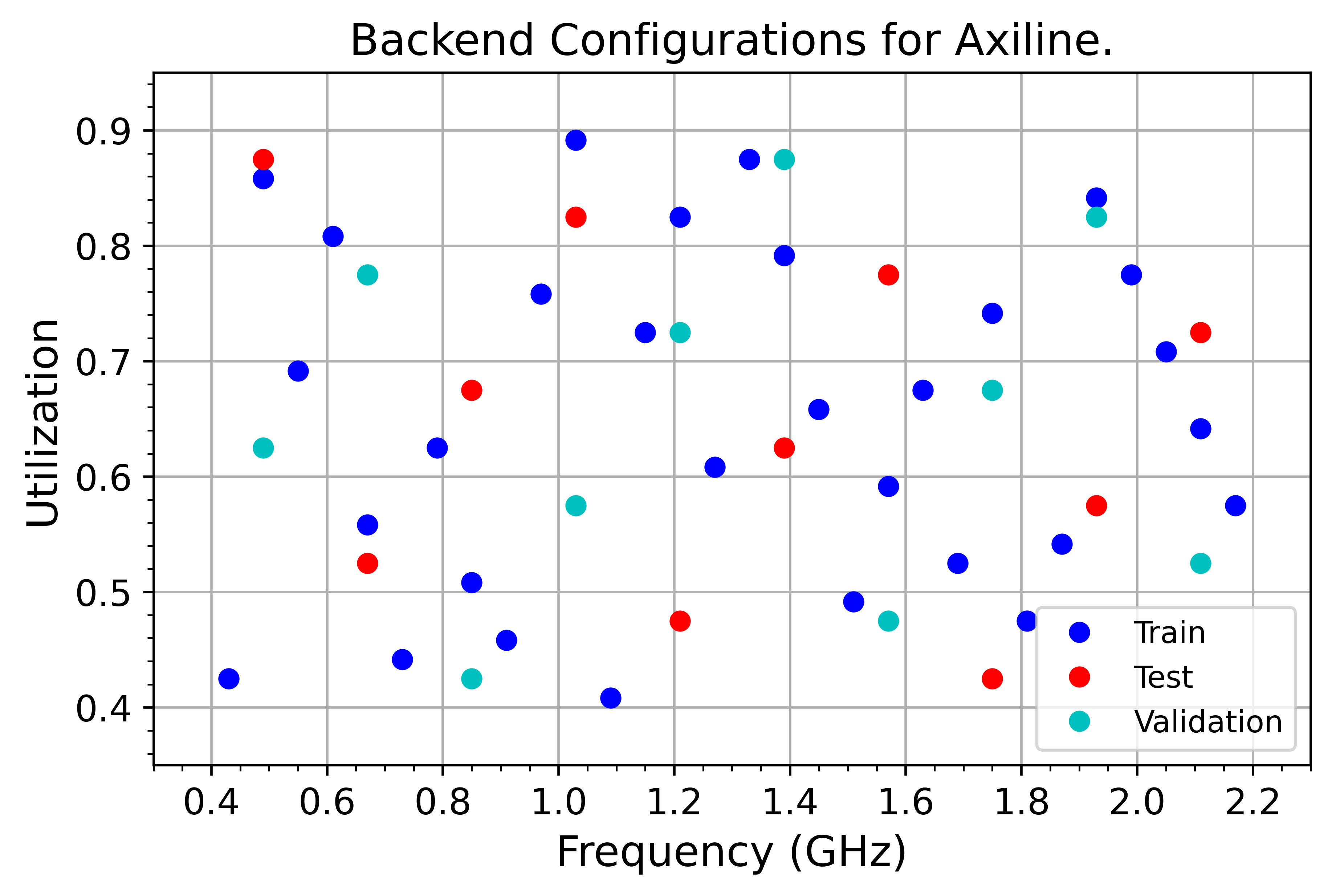}}
    \end{subfigure}
    \begin{subfigure}[]
        {\includegraphics[width=0.48\columnwidth]{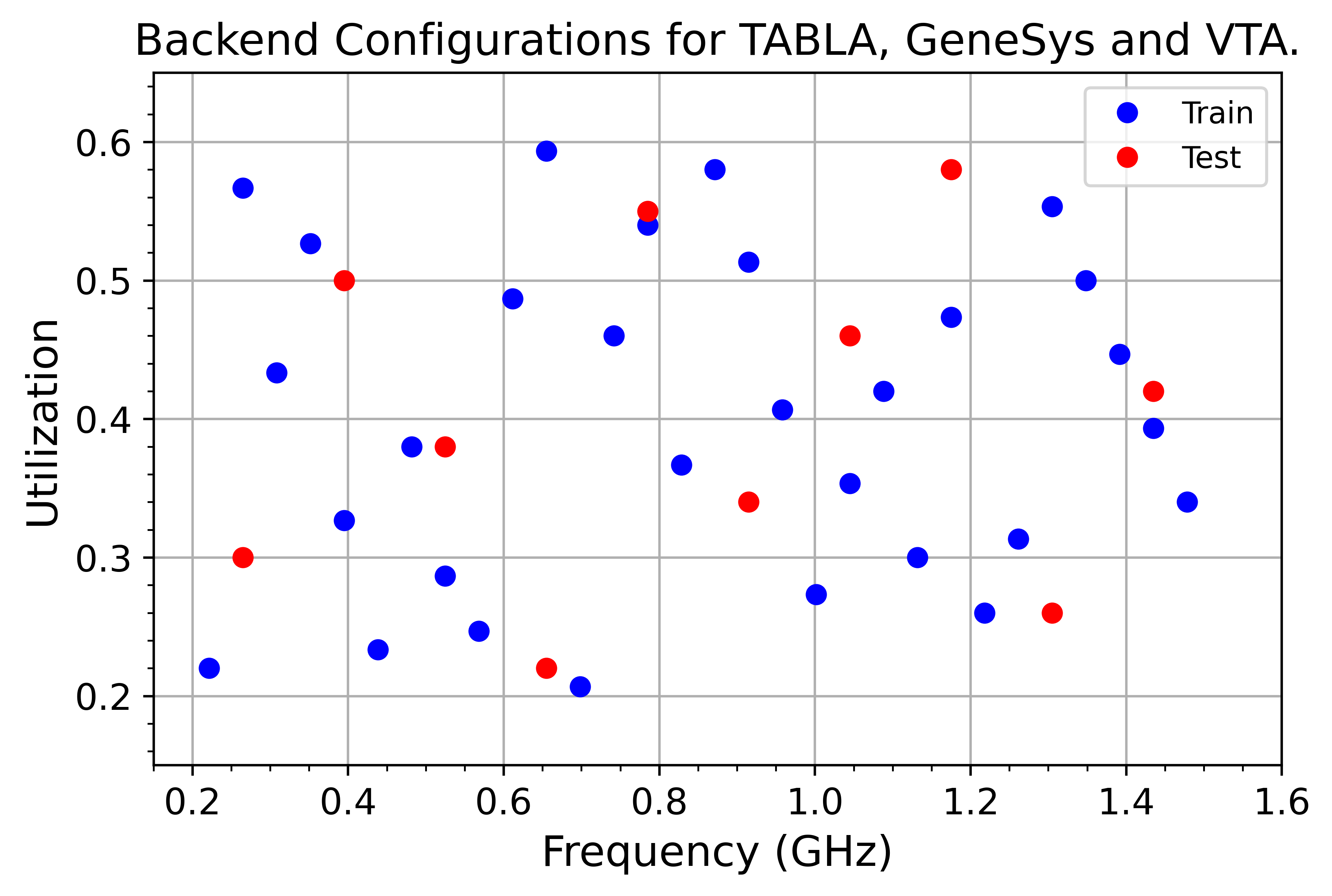}}
    \end{subfigure}
    \caption{\sk{Backend configurations sampled using Latin hypercube sampling.
    Blue dots correspond to the training dataset and
    red dots correspond to the testing dataset.}}
    \label{fig:backend_configurations}
\end{figure}
\noindent

\noindent
\sk{{\bf Sampling of backend parameters.} As mentioned in Section~\ref{sec:problem_formulation}, we 
incorporate the floorplan utilization ($util$) as another backend parameter 
alongside the target clock frequency. This allows for the prediction of 
chip area. If we were to separately sample the target clock period and
floorplan utilization, and then enumerate all configurations, it would
significantly inflate the total number of configurations and be highly
inefficient. Thus, we employ LHS to sample backend configurations. 
The performance of a chip is directly proportional to the clock
frequency, while it is inversely proportional to the clock period.
So, we sample from the target clock
frequency space during backend parameter sampling. We then convert
the frequency to the target clock period for 
the SP\&R data generation. 
Figures~\ref{fig:backend_configurations}(a) and (b)
show the backend configurations sampled for Axiline,
and for TABLA, GeneSys
and VTA designs. For Axiline, we sample floorplan utilization from 
40\% to 90\% and target clock frequency from 0.4 GHz to 2.2 GHz. For
macro-heavy TABLA, GeneSys and VTA designs, we sample floorplan
utilization from 20\% to 60\% and frequency from 0.2 GHz to 1.5 GHz.
}
  
\noindent
\sk{
{\bf PPA and system-level metric data generation.} After 
sampling the architectural configurations, generating the RTL,
and sampling the backend configurations, we run commercial
synthesis, place and route, and collect post-route optimization
PPA metrics. We execute the logic
synthesis using Synopsys Design Compiler R-2020.09, generating
the synthesized netlist. We then carry out place and route 
using Cadence Innovus 21.1 to capture the post-route
optimization PPA metrics. All of our tool scripts are available
in the VeriGood-ML GitHub repository~\cite{veriGoodMLRepo}.
For macro-heavy designs, we employ Innovus' concurrent macro
placer to automatically place all the macros. We conduct all of
our studies using the GLOBALFOUNDRIES 12LP (GF12) enablement. For
Axiline, we also generate PPA metrics on the open enablement
NanGate45~\cite{ng45} to demonstrate the adaptability of our
framework across different platforms. Once we have the
post-route optimization database, we run simulations to
capture the system-level runtime and energy metrics.}
Here, we set the workloads to be the widely-used ResNet-50 and MobileNet-v1
networks for the GeneSys and VTA designs,
respectively. The workload of each TABLA or Axiline design
is determined by its benchmark parameter 
(see Table~\ref{tab:features}).

\subsection{Dataset Separation}
\label{subsec:data_separataion}
\sk{
After generating the data, we separate it into
training and testing datasets. We train and test our
model using two types of datasets, specifically divided
based on backend and architectural configurations.
These are respectively referred to as the {\em unseen
backend dataset} and the {\em unseen architectural
dataset}, respectively.}

\noindent
\sk{{\bf Unseen backend dataset.} As the name implies, the 
training and testing datasets encompass entirely different
sets of backend configurations, though the architectural
configurations within both datasets remain the same. We
sample 30 data points for training and 10 for testing, 
making sure there is no overlap
between any of these datasets. This distinct sampling
helps to
ensure coverage of the entire backend
design space. Figure~\ref{fig:backend_configurations} 
displays the backend configurations sampled for Axiline,
TABLA, GeneSys and VTA for training, and
testing with GF12 enablement. For Axiline, an additional
10 data points are sampled for model validation during
the training process.}

\noindent
\sk{{\bf Unseen architectural dataset.} In this case, 
the training and testing datasets comprise entirely
of different sets of architectural configurations, 
while the backend configurations remain consistent
across both datasets. Just as with unseen backend
configurations, we ensure no overlap between the
training and testing datasets. For the Axiline
design, the training dataset contains 24
configurations, while the validation and testing
datasets each consist of 10 configurations.
Again, we separately sample all three
(i.e., training, validation, and testing)
datasets using LHS to help ensure
a coverage of each 
dataset over the architectural design space.
(Section~\ref{subsec:asses_sampling_methods}
discusses in detail the impact of sampling
training configurations using different methods
for varying sample sizes.) For TABLA, GeneSys,
and VTA, we randomly divide the dataset into
a 4:1 ratio for training and testing,
respectively, based on architectural
configurations.}

\sk{In contrast to our previous work~\cite{Esmaeilzadeh22},
where we randomly divided the training and testing
datasets in a 4:1 ratio, in this study we separately
sample training and testing configurations. Given
the better coverage of the testing dataset, effective
performance of the trained model on these datasets
is a stronger indicator of promising
performance on unseen datasets during DSE.}

\subsection{Model Training}
\label{subsec:model-training}
\sk{
In this work, we train and evaluate a total of 200 ML models,
spanning four platforms (Axiline, TABLA, GeneSys and
VTA), five metrics (power, performance, area, energy, and
runtime), five types of machine learning models (GBDT, RF,
ANN, Stacked Ensemble and GCN) and two datasets (unseen
backend and unseen architectural). We tune the
hyperparameters of each model except GCN, using
the H2O package and a random discrete search method.
Table~\ref{tab:parameters} presents the hyperparameters
of each machine learning model type that we tune. In 
the following subsections, we first discuss the
importance of having a separate validation dataset
instead of relying solely on cross-validation, and then
delve into the detailed steps of hyperparameter tuning
for each model.}
  
\noindent
\sk{{\bf Preference for validation dataset over cross-validation.}
We find that one of the primary causes of the
poor performance of the ANN model in our previous study~\cite{Esmaeilzadeh22}
is related to the use of cross-validation. Using 
cross-validation does not necessarily ensure adequate coverage
of the design space, despite the fact that the complete 
training dataset has this property. This leads to a bias in the model 
that develops during the training process, which results in high
$\mu$APE and MAPE. As mentioned in 
Section~\ref{subsec:data_separataion}, by separately
sampling a validation dataset that provides a thorough
coverage of the design space, we achieve a substantial
improvement in the performance of the model on the testing
dataset. During hyperparameter tuning, we employ the
validation dataset to select the best hyperparameters
and then deploy them on the testing dataset for evaluation.
We apply this approach to the model training of
Axiline designs. However, for TABLA, GeneSys, and VTA 
designs, generating the RTL netlist for different
architectural configurations requires significant
manual effort. So, we opt for five-fold
cross-validation for these designs.}

\noindent
\sk{{\bf Training of GBDT and RF.} We use H2O and adopt
the same strategy to train both the GBDT and RF models.
Hyperparameter tuning using H2O random 
discrete~\cite{RandomDescrete} grid search
is executed in two stages. In
the first stage, we set the number of trees to a 
very large value (500 for RF and 300 for XGB) and
tune the remaining hyperparameters. From the best
hyperparameter configuration, we find the best
$max\_depth$ to narrow down the search space for
$max\_depth$. For RF, we also reduce the search
space for $mtries$. The range for $max\_depth$
is reduced to best $max\_depth \pm 3$. For RF,
we retain the $mtries$ value determined
in the first stage. In the second stage, we
conduct another round of hyperparameter search
with the updated search space for $max\_depth$
in both RF and GBDT and for $mtries$ in RF.
We choose the model with the lowest Root Mean 
Square Error (RMSE), computed using Equation~\eqref{eq:RMSE}
on the validation dataset during the random 
discrete search process. This optimized model
is then applied to the testing dataset.

\begin{equation}
\label{eq:RMSE}
    RMSE = \sqrt{\frac{1}{n} \sum_{i=1}^{n} (y_{actual} - y_{predicted})^2}
\end{equation}
}

\begin{table}[htbp]
    \centering
    \caption{Tuned hyperparameters for {\em GBDT}, {\em RF} and {\em ANN}.}
    \label{tab:parameters}
    \resizebox{\columnwidth}{!}{
    \begin{tabular}{|l|l|l|l|l|}
    \hline
    \textbf{Model} & \textbf{Parameters} & \textbf{Type} 
    & \textbf{Range} & \textbf{Description} \\ \hline
    \multirow{2}{*}{\em GBDT} & n\_estimator & integer 
    & [20 -- 500] & \# gradient boosted trees \\ \cline{2-5}
    & max\_depth & integer 
    & [2 -- 20] & maximum tree depth \\ \cline{2-5} \hline
    \multirow{3}{*}{\em RF} & n\_estimator & integer 
    & [50 -- 1000] & \# decision trees in the forest  \\ \cline{2-5}
    &  mtries & $enum$ & [1 -- total feature count] 
    & \# features considered for best split \\ \cline{2-5}
    &  max\_depth & integer & [5 -- 100] 
    & max tree depth \\ \cline{2-5} \hline
    \multirow{3}{*}{\em ANN} & num\_layer & integer 
    & [3 -- 9] & \# hidden layers \\ \cline{2-5}
    & num\_node & enum & [8, 16, 32] 
    & $nodeCount$ input in Algorithm~\ref{algo:hlayer_config} \\ \cline{2-5}
    & act\_func & {\em enum} & [Tanh, Rectifier, Maxout] 
    & activation function \\ \hline
    \multirow{5}{*}{\em GCN}
    & conv\_layer & {\em enum} & [GraphConv, GCNConv]
    & type of graph convolutional layer \\ \cline{2-5}
    & num\_conv\_layer  & integer  & [2 -- 6]  
    & \# convolutional layers  \\ \cline{2-5}
    & num\_fc\_layer  & integer   & [2 -- 9]  & \# fully connected layers \\ \cline{2-5}
    & batch\_size & integer & [16, 32, 64]
    & training batch size \\ \cline{2-5}
    & lr & float & [$10^{-2} - 10^{-5}$] & learning rate \\ \hline
    \end{tabular}
    }
\end{table} 

\begin{algorithm}[htbp]
\caption{\sk{Generation of hidden layer configurations.}}
\label{algo:hlayer_config}
\SetAlgoLined
\KwIn{nodeCount, hLayerCount, minP = 2, maxP = 7}
\KwOut{$layer$ -- $i^{th}$ element of the list is
the number of nodes in the $i^{th}$ hidden layer.}
\SetKwFunction{FMain}{getNodeConfig}
\SetKwProg{Fn}{Function}{:}{}
\Fn{\FMain{nodeCount, hLayerCount, minP, maxP}}{
    $P = ceil(log_2(nodeCount)$\;
    $expMaxP = min((hLayerCount + minP + P) / 2, maxP)$\;
    \If{expMaxP $\leq$ P}{
        $expMaxP = P + 1$\;
    }
    $incrP = expMaxP - P$\;
    $decrP = min(expMaxP - minP + 1, hLayerCount - incrP)$\;
    $sameP = 0$\;
    \If{$hLayerCount > incrP + decrP$}{
        $sameP = hLayerCount - incrP - decrP$\;
    }
    $layer = []$\;
    Add $2^P$ to $layer$ $incrP$ times while increasing $P$ by $1$\;
    Add $2^P$ to $layer$ $sameP$ times\;
    Add $2^P$ to $layer$ $decrP$ times while decreasing $P$ by $1$\;
    \KwRet{$layer$}\;
}
\end{algorithm}
  
\sk{
\noindent
{\bf Training of ANN.}
For the ANN model, determining an effective hidden layer
configuration is crucial. In previous work~\cite{Esmaeilzadeh22},
hidden layer configurations are generated such that all hidden
layers have identical node configurations, which has further
room for improvement.
The key idea is to map the features to a higher dimensional
space and then gradually reduce them to a smaller dimension.
This allows the model to extract complex representations and
make the data more separable. To address this, we apply
Algorithm~\ref{algo:hlayer_config} to create the hidden layer
configurations. In Algorithm~\ref{algo:hlayer_config},
$nodeCount$ represents the count of
nodes in the first hidden layer, and $hLayerCount$ denotes
the total number of hidden layers. The values $minP$
and $maxP$ are used to constrain the minimum and maximum number
of nodes in the hidden layer, respectively, to $2^{minP}$ and
$2^{maxP}$.}
\sk{
\begin{itemize}[noitemsep,topsep=0pt,leftmargin=*]
    \item In lines 2-3, we calculate $expMaxP$. The value 
    $2^{expMaxP}$ represents the expected maximum number 
    of nodes in the hidden layer. 
    \item In lines 4-6, we update $expMaxP$ if the expected
    maximum node count is fewer than the number of input nodes.
    \item In lines 7-12, we determine respectively the number
    of layers where the node count will increase, decrease,
    and remain constant at $2^{expMaxP}$.
    \item In lines 13-17, we calculate the node
    count for each layer and return the layer configuration.
\end{itemize}
}

\sk{Our algorithm consistently uses a power-of-two
for the node count in each hidden layer, which provides high
resource-efficiency. We employ the parameter values provided in
Table~\ref{tab:parameters} to generate all the configurations
for the hidden layers. We perform the hyperparameter search
for these hidden layer configurations and the set of
activation functions presented in Table~\ref{tab:parameters}.
During model training, we leverage adaptive learning
rates. The model exhibiting the lowest RMSE value 
(Equation~\eqref{eq:RMSE}) on the validation dataset is
selected. We carry out this hyperparameter tuning of ANN
models using the H2O random discrete~\cite{RandomDescrete}
grid search. The selected model is then
applied to the testing dataset for evaluation.}

\noindent
\sk{{\bf Training of Stacked Ensemble.}
We employ the stacked ensemble model of H2O, where we use
the trained models of GBDT, RF and ANN as base learners,
with linear regression acting as meta learner. Here, only
the top seven models from the hyperparameter search process
of GBDT, RF and ANN are chosen as the base
learners.\footnote{Our background studies empirically
suggest that choosing the top seven models gives the best results.}
This approach ensures a sufficient degree of variation
in the selection of base learners,
while filtering out the poorly performing models.}

\begin{figure}
    \centering
    \includegraphics[width = \columnwidth]{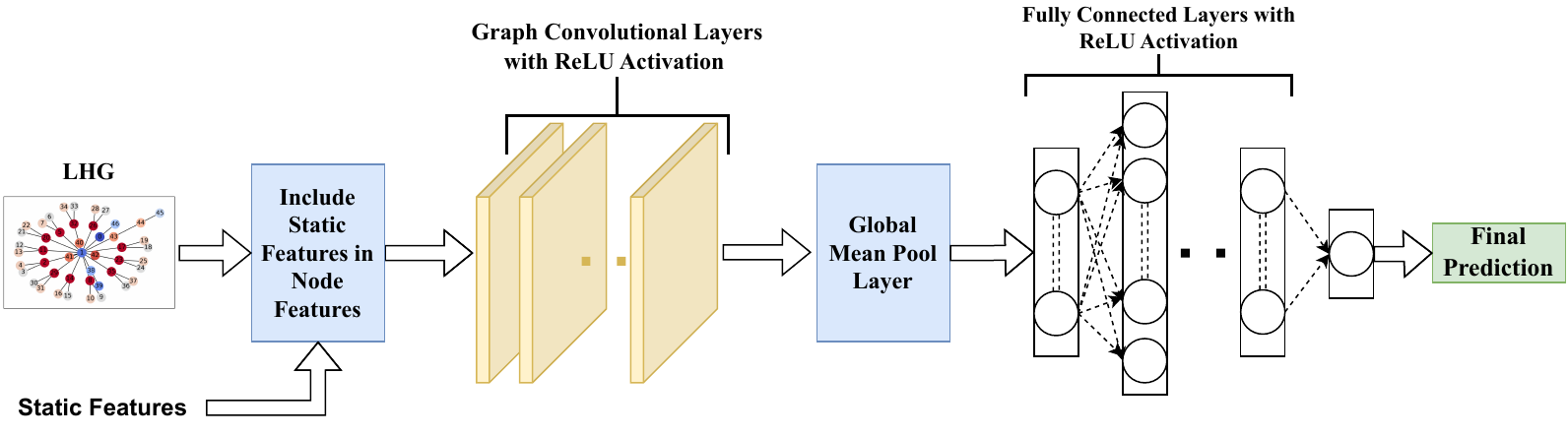}
    \caption{\sk{Configuration of the GCN architecture used for backend PPA
    and system-metric prediction.}}
    \label{fig:gcn_architecture}
\end{figure}

\noindent
\sk{{\bf Training of GCN.} We have implemented the GCN model using 
PyTorch and PyTorch Geometric. The GCN architecture used to train
our model is depicted in Figure~\ref{fig:gcn_architecture}. For 
both the graph convolutional layers and the fully connected 
layers, we use the $ReLU$ activation function. The convolutional
layers generate node embeddings, and we then employ the 
$GlobalMeanPool$ function that computes the average of all the node embeddings, i.e.,
\begin{equation}
\label{eq:global_mean_pool}
    GlobalMeanPool(\mathbf{X}) = \frac{1}{N} \sum_{i=1}^{N} \mathbf{X}_i
\end{equation}
}

\begin{figure}[htbp]
    \centering
    \begin{subfigure}[]
    {\includegraphics[width = 0.31\columnwidth]{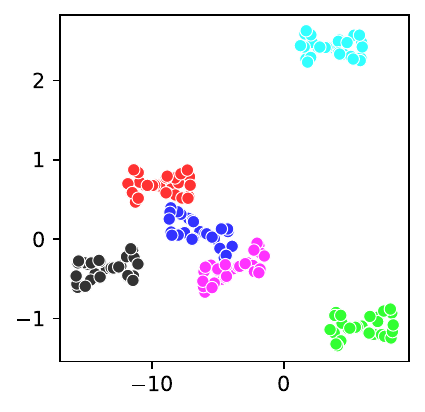}}
    \end{subfigure}
    \begin{subfigure}[]
    {\includegraphics[width = 0.335\columnwidth]{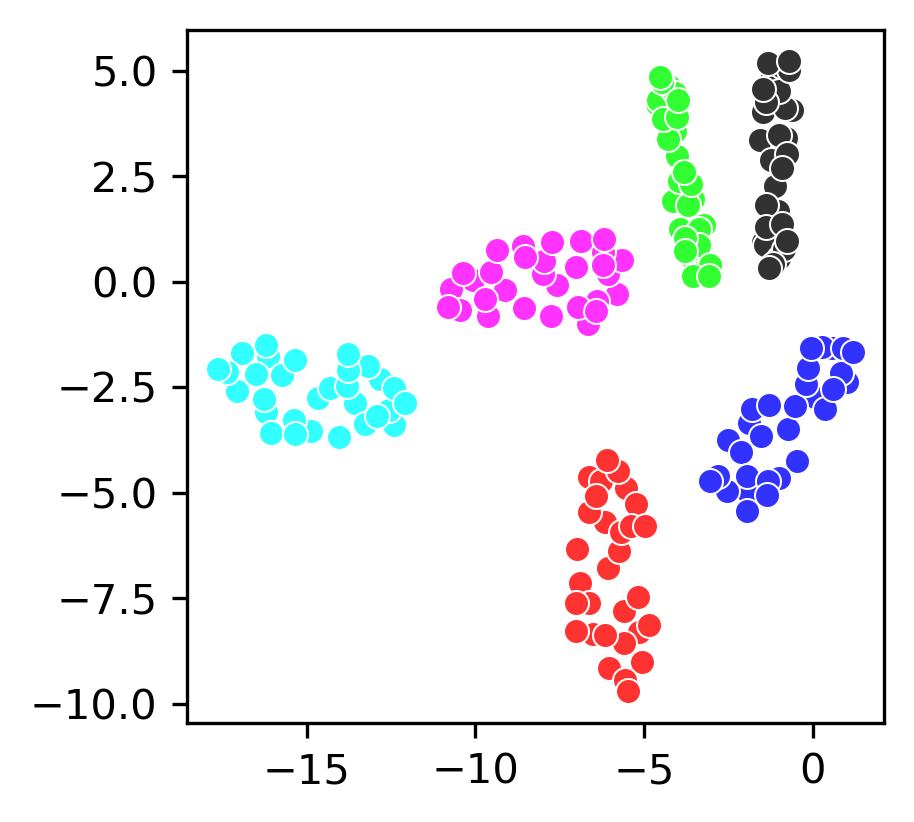}}
    \end{subfigure}
    \begin{subfigure}[]
    {\includegraphics[width = 0.32\columnwidth]{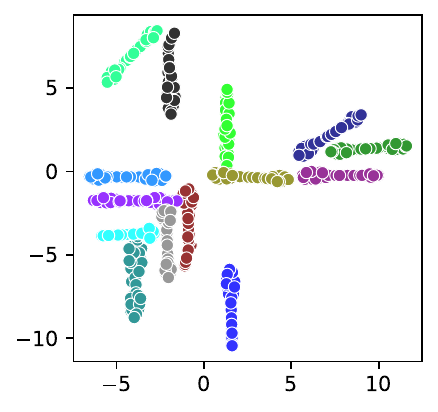}}
    \end{subfigure}
    \caption{\sk{t-SNE plot of the graph embedding generated
    by the trained GCN models for (a) TABLA, (b) VTA and
    (c) Axiline designs. Here, the data points highlighted
    with same color represent the same architectural
    configuration, but with different backend
    configurations.}}
    \label{fig:graph_embedding}
\end{figure}

\sk{The GCN models incorporate not only architectural and
backend features but also extract features from $LHG$s. Given that
ML accelerators are inherently modular and have shared building
blocks, this enables the GCN models to produce more insightful
embeddings for PPA and system-level metric predictions.
Figure~\ref{fig:graph_embedding} shows the t-SNE plot of the graph
embeddings generated for TABLA, VTA, and Axiline designs. Here,
different colors are used to plot graph embeddings of different
architectural configurations. In the t-SNE plot, we see clear
distinctions between different architectural configurations.
Since the PPA and system-level metrics differ for these
configurations, it indicates that the GCN models are
well-trained.}

\sk{Following the generation of graph embeddings, we input 
them into the fully connected layer to produce the final
predictions. The configuration of the fully connected layer
is generated using the $getNodeConfig$ function, with the graph
embedding size and the number of fully connected layers serving
as inputs, corresponding to $nodeCount$ and $hLayerCount$,
respectively. While training our model, we employ the 
mean absolute percentage error ($\mu APE$) loss, defined as
\begin{equation}
\label{eq:mape_loss}
    \mu APE = \frac{1}{n} \sum_{i=1}^{n} \left|\frac{{y_{actual} - y_{predicted}}}{{y_{actual}}}\right| \times 100
\end{equation}
}

\sk{Throughout the training of the GCN model, we utilize the Adam~\cite{Adam}
optimizer with decaying learning rate; decaying factor
is set to 0.7 with a patience of 5. We also implement early
stopping if no improvement in the model's performance on the
validation dataset is observed for 20 consecutive epochs.
Upon completion of the training, the model exhibiting the
lowest validation error is chosen for testing on the test
dataset.}

\sk{We employ the {\em HyperOptSearch} function of
Ray~Tune~\cite{RayTune} to automatically
optimize various parameters such as the type of graph
convolutional layer, the number of convolutional layers,
the number of fully connected layers, the batch size,
and the learning rate as shown in Table~\ref{tab:parameters}.
The best configuration is determined
based on the following loss function,\footnote{In our
experiments with the GCN model, the value of $\mu APE$
varies within the range [0 -- 20], whereas $MAPE$ varies
in the range [0 -- 60]. To ensure both $\mu APE$ and
$MAPE$ on the same scale, we use a weight of 
0.3 ($20/60 \approx 0.3$) for $MAPE$.}
which captures both the average and worst-case performance: 
\begin{equation}
\label{eq:tune_loss_function}
   loss = {\mu APE} + 0.3\times{MAPE}
\end{equation}
}

\section{Experimental results}
\label{sec:experimental_results}

\sk{In this section, we present our experimental
results. First, we show the performance of
our model using various sampling methods and
sample sizes. We then demonstrate the
performance of the model for unseen backend
and architectural configurations. Next,
we study our model performance for limited training
dataset. Finally, we utilize the trained model for
the DSE of Axiline and VTA designs.}

\begin{figure}
    \centering
    \begin{subfigure}[]
        {\includegraphics[width=0.45\columnwidth]{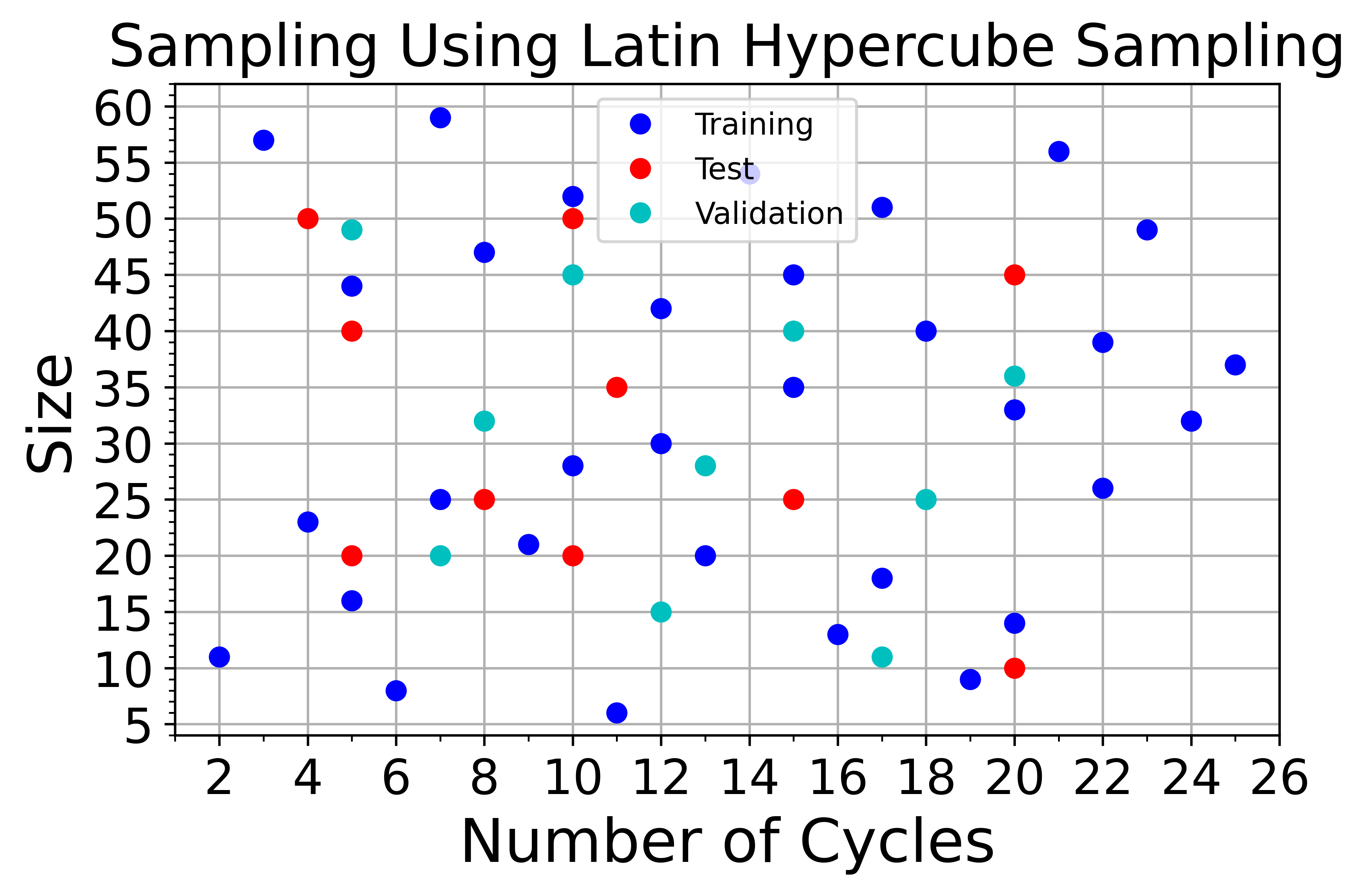}}
    \end{subfigure}
    \begin{subfigure}[]
        {\includegraphics[width=0.45\columnwidth]{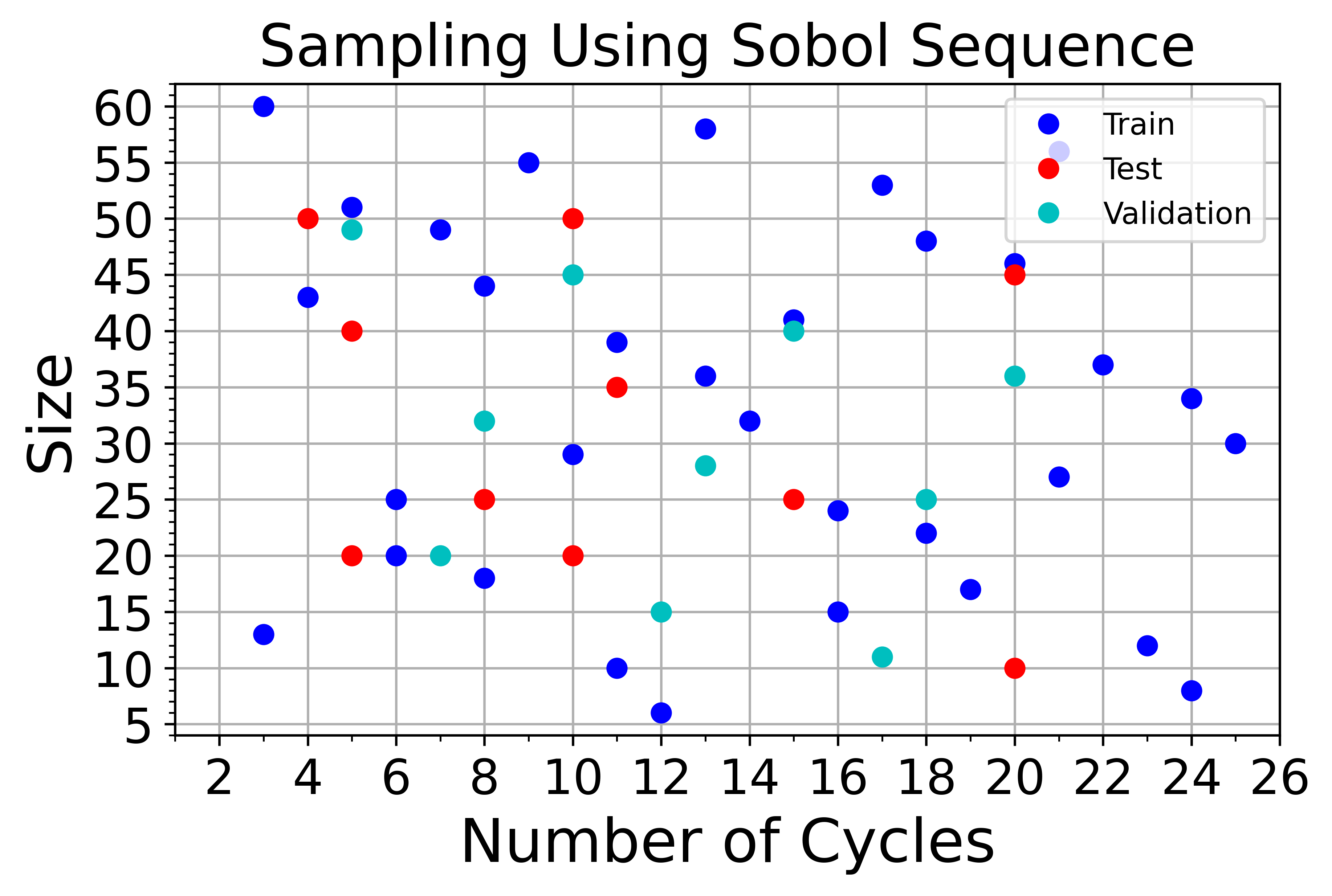}}
    \end{subfigure}
    \begin{subfigure}[]
        {\includegraphics[width=0.45\columnwidth]{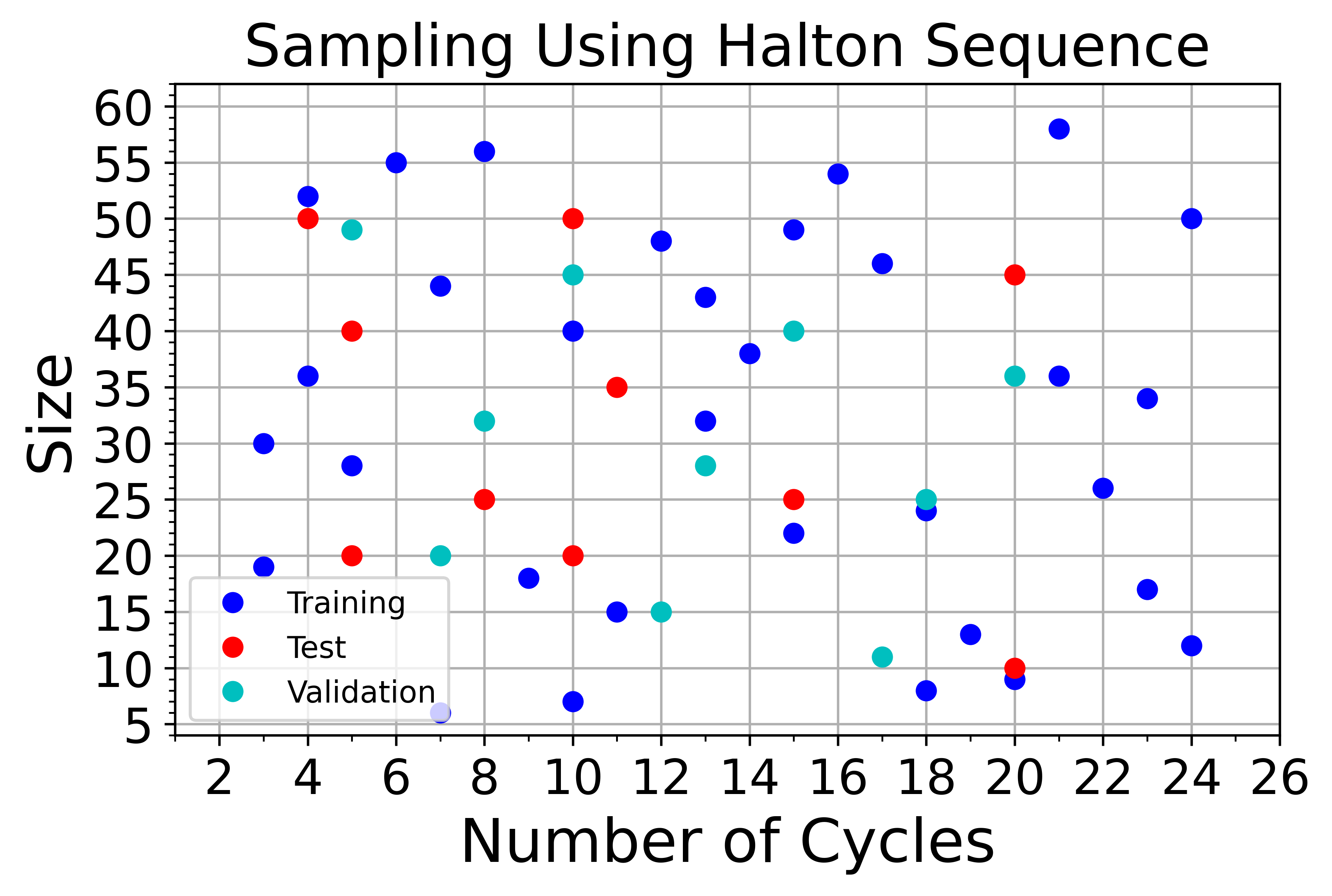}}
    \end{subfigure}
    \caption{\sk{Axiline architectural configurations sampled using
    (a) Latin Hypercube Sampling, (b) Sobol Sequence, and 
    (c) Halton Sequence. Blue, cyan, and red dots respectively
     denote training, validation, and testing configurations.}}
    \label{fig:sampling_method}
\end{figure}

\subsection{Assessment of Sampling Methods and Sample Sizes}
\label{subsec:asses_sampling_methods}
\sk{The selection of an appropriate sampling method and sample
size is crucial, as a poor choice may result in non-uniform 
sampling. This can subsequently lead to the development of 
biased models or necessitate a larger volume of data points 
in order to achieve the desired performance. To identify the
appropriate sampling method and sample size, we train our
models using data obtained from various sampling techniques
and a range of sample sizes. Subsequently, we assess the
performance of the trained models on unseen test
configurations, capturing the mean absolute percentage error
($\mu APE$) and maximum absolute percentage error ($MAPE$).
Additionally, we compute standard deviation of $APE$ (STD $APE$)
on the test configurations to measure stability of model
performance on unseen configuration across the architectural
parameter space. A smaller value of STD $APE$ indicates
stable model performance on the unseen configurations.}

\sk{Table~\ref{tab:sampling_method_size} shows the performance
of different ML models for Axiline-SVM on the testing dataset when the
training data is sampled using Latin Hypercube Sampling, Sobol
Sequence, and Halton Sequence with sample sizes of 16, 24, and 32.
For a given sample size and machine learning model, we use bold
font to indicate the result of the top-performing sampling method.
From Table~\ref{tab:sampling_method_size}, we make the following
observations.
\begin{itemize}[noitemsep,topsep=0pt,leftmargin=*]
    \item Consistent with expectations, across all sampling methods,
    an increase in sample size leads to a decrease in both $\mu APE$
    and STD $APE$, thus indicating an improvement in model 
    performance.
    \item When considering smaller sample sizes, the GCN models 
    demonstrate superior results in terms of both STD $APE$
    and $MAPE$ in comparison to other models.
    \item In 12/24 instances, LHS yields superior results in terms
    of $\mu APE$, and in 10/24 instances, in terms of both $MAPE$
    and STD $APE$, in comparison to other sampling techniques.
    Overall, LHS demonstrates better performance than
    either the Sobol or Halton methods. Consequently, we choose to
    use LHS in experiments reported below.
    \item The model's performance, across all metrics, does not
    exhibit substantial improvements when the sample size is
    increased from 24 to 32. Hence, for the experiments
    reported below, we maintain a sample size of 24.
\end{itemize} 
\noindent
To recap: in the remaining portion of our experiments, we employ
LHS with a sample size of 24 to generate the training dataset for Axiline
designs.
}

\begin{table}[htbp]
\customsize
\centering
\caption{\sk{ML model performance on unseen architectural configurations
for different sampling methods and sample sizes. Here, the standard
deviation of $APE$ (STD $APE$) represents the variation in $APE$ across
all different test configurations. A smaller value of STD $APE$ 
indicates stable performance on unseen architectural configurations.}}
\label{tab:sampling_method_size}
\begin{tabular}{|cc|l|llr|llr|}
\hline
\multicolumn{2}{|c|}{Sampling Details}                               
& \multicolumn{1}{c|}{\multirow{2}{*}{ML Model}} 
& \multicolumn{3}{c|}{Backend Power}                                    
& \multicolumn{3}{c|}{System-Energy}          \\ \cline{1-2} \cline{4-9} 
\multicolumn{1}{|c|}{Method}                   
& Size                
& \multicolumn{1}{c|}{}                          
& \multicolumn{1}{l|}{$\mu APE$} & \multicolumn{1}{l|}{STD $APE$} 
& $MAPE$ & \multicolumn{1}{l|}{$\mu APE$} 
& \multicolumn{1}{l|}{STD $APE$} & $MAPE$ \\ \hline
\multicolumn{1}{|c|}{\multirow{12}{*}{LHS}}    
& \multirow{4}{*}{16} 
& GBDT                                           
& \multicolumn{1}{r|}{20.37}          
& \multicolumn{1}{r|}{\bf 13.14}    & 84.16
& \multicolumn{1}{r|}{32.65}          
& \multicolumn{1}{r|}{25.68}    & 88.69      \\ \cline{3-9} 
\multicolumn{1}{|c|}{}                         & 
& RF                                             
& \multicolumn{1}{r|}{\bf 17.63}
& \multicolumn{1}{r|}{\bf 10.01}   & {\bf 57.00}
& \multicolumn{1}{r|}{\bf 36.58}
& \multicolumn{1}{r|}{23.28}    & {\bf 78.84}     \\ \cline{3-9} 
\multicolumn{1}{|c|}{}                         & 
& ANN                                            
& \multicolumn{1}{r|}{\bf 2.67}          
& \multicolumn{1}{r|}{\bf 1.31}         & 23.02
& \multicolumn{1}{r|}{\bf 4.36}          
& \multicolumn{1}{r|}{2.95}     & 28.13     \\ \cline{3-9} 
\multicolumn{1}{|c|}{}                         & 
& GCN                                            
& \multicolumn{1}{r|}{\bf 2.96}          
& \multicolumn{1}{r|}{\bf 0.45}         & {\bf 15.89}
& \multicolumn{1}{r|}{\bf 3.06}          
& \multicolumn{1}{r|}{0.99}         & {\bf 13.92}     \\ \cline{2-9} 
\multicolumn{1}{|c|}{}
& \multirow{4}{*}{24}
& GBDT                            
& \multicolumn{1}{r|}{13.20}
& \multicolumn{1}{r|}{8.07}         & {\bf 58.63}
& \multicolumn{1}{r|}{\bf 15.25}          
& \multicolumn{1}{r|}{\bf 9.92}         & {\bf 65.02}     \\ \cline{3-9} 
\multicolumn{1}{|c|}{}                         & 
& RF                                             
& \multicolumn{1}{r|}{14.76}          
& \multicolumn{1}{r|}{12.29}         & 84.38
& \multicolumn{1}{r|}{\bf 14.38}          
& \multicolumn{1}{r|}{12.07}         & 68.17     \\ \cline{3-9} 
\multicolumn{1}{|c|}{}                         & 
& ANN
& \multicolumn{1}{r|}{\bf 1.80}          
& \multicolumn{1}{r|}{\bf 0.54}         & 15.83
& \multicolumn{1}{r|}{3.44}          
& \multicolumn{1}{r|}{\bf 2.26}         & {\bf 21.73}     \\ \cline{3-9} 
\multicolumn{1}{|c|}{}                         & 
& GCN
& \multicolumn{1}{r|}{3.00}
& \multicolumn{1}{r|}{0.91}         & {\bf 15.38}
& \multicolumn{1}{r|}{2.71}
& \multicolumn{1}{r|}{0.74}         & 16.21     \\ \cline{2-9}
\multicolumn{1}{|c|}{}
& \multirow{4}{*}{32}
& GBDT
& \multicolumn{1}{r|}{13.61}
& \multicolumn{1}{r|}{6.07}         & 59.70
& \multicolumn{1}{r|}{\bf 9.75}
& \multicolumn{1}{r|}{\bf 6.69}         & 48.34     \\ \cline{3-9} 
\multicolumn{1}{|c|}{}                         & 
& RF
& \multicolumn{1}{r|}{12.06}
& \multicolumn{1}{r|}{\bf 6.73}         & 44.10
& \multicolumn{1}{r|}{21.88}
& \multicolumn{1}{r|}{20.00}         & 84.68     \\ \cline{3-9} 
\multicolumn{1}{|c|}{}                         & 
& ANN
& \multicolumn{1}{r|}{\bf 2.03}
& \multicolumn{1}{r|}{0.70}         & {\bf 13.24}
& \multicolumn{1}{r|}{4.00}          
& \multicolumn{1}{r|}{3.65}         & 31.51 \\ \cline{3-9} 
\multicolumn{1}{|c|}{}                         & 
& GCN
& \multicolumn{1}{r|}{\bf 2.57}          
& \multicolumn{1}{r|}{0.70}         & {\bf 15.99}
& \multicolumn{1}{r|}{2.20}          
& \multicolumn{1}{r|}{\bf 0.66}         & 19.92     \\ \hline
\multicolumn{1}{|c|}{\multirow{12}{*}{Sobol}}
& \multirow{4}{*}{16}
& GBDT                                           
& \multicolumn{1}{r|}{\bf 18.02}          
& \multicolumn{1}{r|}{15.64}     & {\bf 72.16}
& \multicolumn{1}{r|}{\bf 28.19}          
& \multicolumn{1}{r|}{\bf 16.12}    & {\bf 82.50}     \\ \cline{3-9} 
\multicolumn{1}{|c|}{}                         & 
& RF                                             
& \multicolumn{1}{r|}{22.58}          
& \multicolumn{1}{r|}{15.07}         & 75.50
& \multicolumn{1}{r|}{39.63}          
& \multicolumn{1}{r|}{\bf 15.15}         &  85.97    \\ \cline{3-9} 
\multicolumn{1}{|c|}{}                         & 
& ANN                                            
& \multicolumn{1}{r|}{3.18}          
& \multicolumn{1}{r|}{1.52}     & {\bf 20.67}
& \multicolumn{1}{r|}{5.16}          
& \multicolumn{1}{r|}{\bf 2.15}     & {\bf 24.11}     \\ \cline{3-9} 
\multicolumn{1}{|c|}{}                         & 
& GCN                                            
& \multicolumn{1}{r|}{3.24}          
& \multicolumn{1}{r|}{0.94}         & 18.03
& \multicolumn{1}{r|}{3.32}
& \multicolumn{1}{r|}{\bf 0.87}         & 22.39     \\ \cline{2-9} 
\multicolumn{1}{|c|}{}                         
& \multirow{4}{*}{24} 
& GBDT                                           
& \multicolumn{1}{r|}{14.31}          
& \multicolumn{1}{r|}{10.65}    & 80.92
& \multicolumn{1}{r|}{34.14}          
& \multicolumn{1}{r|}{33.93}    & 99.20     \\ \cline{3-9} 
\multicolumn{1}{|c|}{}                         & 
& RF                                             
& \multicolumn{1}{r|}{18.32}          
& \multicolumn{1}{r|}{13.78}         & 73.92
& \multicolumn{1}{r|}{29.41}          
& \multicolumn{1}{r|}{19.15}         & 89.63     \\ \cline{3-9} 
\multicolumn{1}{|c|}{}                         & 
& ANN                                            
& \multicolumn{1}{r|}{2.70}          
& \multicolumn{1}{r|}{1.31}     & 27.40
& \multicolumn{1}{r|}{5.19}          
& \multicolumn{1}{r|}{2.45}    & 22.21     \\ \cline{3-9} 
\multicolumn{1}{|c|}{}                         & 
& GCN                                            
& \multicolumn{1}{r|}{\bf 2.51}          
& \multicolumn{1}{r|}{1.05}       & 15.89
& \multicolumn{1}{r|}{2.62}
& \multicolumn{1}{r|}{\bf 0.69}       & {\bf 15.85}  \\ \cline{2-9} 
\multicolumn{1}{|c|}{}                         
& \multirow{4}{*}{32} 
& GBDT                                           
& \multicolumn{1}{r|}{14.95}          
& \multicolumn{1}{r|}{9.98}    & 37.90
& \multicolumn{1}{r|}{21.45}          
& \multicolumn{1}{r|}{16.52}    & {\bf 34.89}     \\ \cline{3-9} 
\multicolumn{1}{|c|}{}                         & 
& RF                                             
& \multicolumn{1}{r|}{16.06}          
& \multicolumn{1}{r|}{12.74}         & {\bf 33.85}
& \multicolumn{1}{r|}{25.84}          
& \multicolumn{1}{r|}{27.19}         & {\bf 46.02}     \\ \cline{3-9} 
\multicolumn{1}{|c|}{}                         & 
& ANN                                           
& \multicolumn{1}{r|}{2.39}          
& \multicolumn{1}{r|}{1.00}         & 25.07
& \multicolumn{1}{r|}{2.59}          
& \multicolumn{1}{r|}{1.46}         & 21.31     \\ \cline{3-9} 
\multicolumn{1}{|c|}{}                         & 
& GCN                                            
& \multicolumn{1}{r|}{2.58}          
& \multicolumn{1}{r|}{0.72}         & 18.03
& \multicolumn{1}{r|}{\bf 2.14}          
& \multicolumn{1}{r|}{0.72}         & 18.87     \\ \hline
\multicolumn{1}{|c|}{\multirow{12}{*}{Halton}} 
& \multirow{4}{*}{16} 
& GBDT                                           
& \multicolumn{1}{r|}{19.28}          
& \multicolumn{1}{r|}{15.32}         & 80.90
& \multicolumn{1}{r|}{48.54}          
& \multicolumn{1}{r|}{43.57}         & 85.08    \\ \cline{3-9} 
\multicolumn{1}{|c|}{}                         & 
& RF                                             
& \multicolumn{1}{r|}{21.46}
& \multicolumn{1}{r|}{18.24}         & 88.02
& \multicolumn{1}{r|}{49.52}
& \multicolumn{1}{r|}{79.62}         & 85.01     \\ \cline{3-9} 
\multicolumn{1}{|c|}{}                         & 
& ANN                                            
& \multicolumn{1}{r|}{4.07}          
& \multicolumn{1}{r|}{2.57}         & 37.25
& \multicolumn{1}{r|}{9.22}          
& \multicolumn{1}{r|}{8.23}         & 60.09     \\ \cline{3-9} 
\multicolumn{1}{|c|}{}                         & 
& GCN                                           
& \multicolumn{1}{r|}{3.81}          
& \multicolumn{1}{r|}{1.46}         & 18.38
& \multicolumn{1}{r|}{4.31}         
& \multicolumn{1}{r|}{1.30}         & 18.38     \\ \cline{2-9} 
\multicolumn{1}{|c|}{}                         
& \multirow{4}{*}{24} 
& GBDT                                          
& \multicolumn{1}{r|}{\bf 12.80}         
& \multicolumn{1}{r|}{\bf 7.80}          & 69.04
& \multicolumn{1}{r|}{26.27}         
& \multicolumn{1}{r|}{16.46}         & 79.35     \\ \cline{3-9} 
\multicolumn{1}{|c|}{}                         & 
& RF 
& \multicolumn{1}{r|}{\bf 13.15}
& \multicolumn{1}{r|}{\bf 10.63}         & {\bf 49.89}
& \multicolumn{1}{r|}{20.57}         
& \multicolumn{1}{r|}{\bf 12.01}         & {\bf 61.67}     \\ \cline{3-9} 
\multicolumn{1}{|c|}{}                         & 
& ANN                                            
& \multicolumn{1}{r|}{1.94}         
& \multicolumn{1}{r|}{0.57}         & {\bf 13.31}
& \multicolumn{1}{r|}{\bf 3.01}          
& \multicolumn{1}{r|}{2.50}         & 32.13     \\ \cline{3-9} 
\multicolumn{1}{|c|}{}                         & 
& GCN                                            
& \multicolumn{1}{r|}{2.65}        
& \multicolumn{1}{r|}{\bf 0.40}         & 17.07
& \multicolumn{1}{r|}{\bf 2.51}         
& \multicolumn{1}{r|}{0.71}         & 16.38     \\ \cline{2-9} 
\multicolumn{1}{|c|}{}                         
& \multirow{4}{*}{32} 
& GBDT                                          
& \multicolumn{1}{r|}{\bf 8.88}         
& \multicolumn{1}{r|}{\bf 3.80}         & {\bf 32.64}
& \multicolumn{1}{r|}{27.48}         
& \multicolumn{1}{r|}{24.46}         & 95.76     \\ \cline{3-9} 
\multicolumn{1}{|c|}{}                         & 
& RF                                            
& \multicolumn{1}{r|}{\bf 11.46}
& \multicolumn{1}{r|}{7.65}         &  40.66
& \multicolumn{1}{r|}{\bf 21.48}
& \multicolumn{1}{r|}{\bf 13.39}        &  58.25    \\ \cline{3-9} 
\multicolumn{1}{|c|}{}                         & 
& ANN
& \multicolumn{1}{r|}{2.27}
& \multicolumn{1}{r|}{\bf 0.59}         & 21.03
& \multicolumn{1}{r|}{\bf 2.44}
& \multicolumn{1}{r|}{\bf 1.08}         & {\bf 18.07}     \\ \cline{3-9}
\multicolumn{1}{|c|}{}                         &
& GCN
& \multicolumn{1}{r|}{2.74}
& \multicolumn{1}{r|}{\bf 0.58}         & 19.73
& \multicolumn{1}{r|}{2.71}
& \multicolumn{1}{r|}{0.84}         & {\bf 15.98}     \\ \hline
\end{tabular}
\end{table}
\subsection{ML Model Assessment}
\label{subsec:model_assessment}
\sk{We now present the results of our model's
performance for predicting backend PPA metrics, and system-level
runtime and energy for unseen backend and
architectural configurations. Our performance evaluation
is based on $\mu APE$, $MAPE$ and STD $APE$. The latter helps us
understand how consistently our model performs on the
testing dataset.}

\noindent
{\bf Results for unseen backend configurations.} \sk{
Table~\ref{tab:unseen_backend} presents the performance
of the ML model for unseen backend configurations in
predicting post-SP\&R PPA and system-level metrics for
TABLA, Genesys, VTA, and Axiline designs, implemented
on GF12 and/or NG45. For the ROI classification task,
all models for the GF12 implementation achieve at least
96\% accuracy and an F1 score of 0.97. For the Axiline
NG45 implementation, all models achieve at least 94\%
accuracy and an F1 score of 0.96. These results
demonstrate the excellent performance of our models
in identifying whether data points belong to the ROI.
In Table~\ref{tab:unseen_backend}, the best-performing
model based on $\mu APE$ for each design and each metric
is highlighted in bold. Based on this table and for
these models, we make the following observations.
\begin{itemize}[noitemsep,topsep=0pt,leftmargin=*]
    \item The best-performing ML model achieves a 
    $\mu APE$ of less than 6\% and a $MAPE$
    of less than 30\% for backend PPA prediction.
    For Axiline-NG45, the backend power 
    prediction using the GCN model yields the highest
    $MAPE$. Upon investigating the model, we find that
    the STD $APE$ on the testing dataset is 4.81.
    \item The best-performing ML model achieves 
    a $\mu APE$ of less than 5\% and a 
    $MAPE$ of less than 37\% for system-level
    metric prediction. 
    For Axiline-NG45, the system-level runtime
    prediction using the GBDT model yields the
    highest $MAPE$. Upon investigating the model,
    we find that the STD
    $APE$ on the testing dataset is 3.77.
\end{itemize}
}

\begin{table}[htbp]
\customsize
\centering
\caption{Performance of ML models for unseen backend configurations.}
\label{tab:unseen_backend}
\begin{tabular}{|c|l|ll|ll|ll|ll|ll|}
\hline
\multirow{2}{*}{Design}
& \multicolumn{1}{c|}{\multirow{2}{*}{ML Model}}
& \multicolumn{2}{c|}{Backend-Perf}
& \multicolumn{2}{c|}{Backend-Power}
& \multicolumn{2}{c|}{Backend-Area}
& \multicolumn{2}{c|}{System-Energy}
& \multicolumn{2}{c|}{System-Runtime} \\ \cline{3-12}
    & \multicolumn{1}{c|}{}
    & \multicolumn{1}{c|}{$\mu APE$} & \multicolumn{1}{c|}{$MAPE$}
    & \multicolumn{1}{c|}{$\mu APE$} & \multicolumn{1}{c|}{$MAPE$} 
    & \multicolumn{1}{c|}{$\mu APE$} & \multicolumn{1}{c|}{$MAPE$} 
    & \multicolumn{1}{c|}{$\mu APE$} & \multicolumn{1}{c|}{$MAPE$} 
    & \multicolumn{1}{c|}{$\mu APE$} & \multicolumn{1}{c|}{$MAPE$} \\ \hline
\multirow{5}{*}{\begin{tabular}[c]{@{}c@{}}TABLA\\ GF12\end{tabular}}
& GBDT
& \multicolumn{1}{l|}{3.09}      & 14.02
& \multicolumn{1}{l|}{2.88}      & 11.94
& \multicolumn{1}{l|}{0.78}      & 3.02
& \multicolumn{1}{l|}{1.22}      & 5.15
& \multicolumn{1}{l|}{3.44}      & 13.28 \\ \cline{2-12} 

& RF                                             
& \multicolumn{1}{l|}{6.13}     & 29.43
& \multicolumn{1}{l|}{3.58}     & 12.15
& \multicolumn{1}{l|}{2.59}     & 10.55
& \multicolumn{1}{l|}{1.83}     & 5.08
& \multicolumn{1}{l|}{6.17}     & 20.39  \\ \cline{2-12} 

& ANN
& \multicolumn{1}{l|}{3.21}     & 11.05
& \multicolumn{1}{l|}{2.88}     & 13.36
& \multicolumn{1}{l|}{\bf 0.24}     & 0.93
& \multicolumn{1}{l|}{\bf 0.85}     & 2.73
& \multicolumn{1}{l|}{3.14}     & 17.15  \\ \cline{2-12} 

& Ensemble
& \multicolumn{1}{l|}{2.82}     & 11.00
& \multicolumn{1}{l|}{2.28}     & 9.51
& \multicolumn{1}{l|}{1.25}     & 6.33
& \multicolumn{1}{l|}{0.93}     & 3.53
& \multicolumn{1}{l|}{3.84}     & 14.55   \\ \cline{2-12}

& GCN
& \multicolumn{1}{l|}{\bf 2.75}     & 11.56
& \multicolumn{1}{l|}{\bf 2.18}     & 8.94
& \multicolumn{1}{l|}{0.54}         & 5.64
& \multicolumn{1}{l|}{0.93}         & 5.39
& \multicolumn{1}{l|}{\bf 3.03}     & 11.82  \\ \hline

\multirow{5}{*}{\begin{tabular}[c]{@{}c@{}}GeneSys\\GF12\end{tabular}} 
& GBDT
& \multicolumn{1}{l|}{7.16}     & 22.74
& \multicolumn{1}{l|}{8.57}     & 26.41
& \multicolumn{1}{l|}{3.93}     & 12.82
& \multicolumn{1}{l|}{1.50}     & 7.03
& \multicolumn{1}{l|}{6.76}     & 20.15  \\ \cline{2-12} 

& RF
& \multicolumn{1}{l|}{11.04}     & 23.52
& \multicolumn{1}{l|}{8.48}     & 18.29
& \multicolumn{1}{l|}{3.25}     & 7.46
& \multicolumn{1}{l|}{1.56}     & 7.60
& \multicolumn{1}{l|}{8.99}     & 34.54  \\ \cline{2-12}

& ANN                                            
& \multicolumn{1}{l|}{6.50}     & 15.49
& \multicolumn{1}{l|}{\bf 5.26}     & 17.93
& \multicolumn{1}{l|}{0.84}     & 2.27
& \multicolumn{1}{l|}{2.80}     & 7.80
& \multicolumn{1}{l|}{6.40}     & 18.46  \\ \cline{2-12} 

& Ensemble                                       
& \multicolumn{1}{l|}{8.38}     & 24.36
& \multicolumn{1}{l|}{6.45}     & 22.02
& \multicolumn{1}{l|}{1.00}     & 3.04
& \multicolumn{1}{l|}{\bf 1.80}     & 5.37
& \multicolumn{1}{l|}{6.45}     & 17.86  \\ \cline{2-12} 

& GCN                                            
& \multicolumn{1}{l|}{\bf 6.00}     & 20.59
& \multicolumn{1}{l|}{7.28}     & 15.81
& \multicolumn{1}{l|}{\bf 0.49}     & 1.30
& \multicolumn{1}{l|}{\bf 1.80}     & 5.11
& \multicolumn{1}{l|}{\bf 5.83}     & 15.51  \\ \hline
\multirow{5}{*}{\begin{tabular}[c]{@{}c@{}}VTA\\ GF12\end{tabular}}     
& GBDT                                           
& \multicolumn{1}{l|}{2.75}     & 13.38
& \multicolumn{1}{l|}{2.84}     & 12.75
& \multicolumn{1}{l|}{1.90}     & 13.14
& \multicolumn{1}{l|}{\bf 1.89}     & 8.27
& \multicolumn{1}{l|}{2.84}     & 12.18  \\ \cline{2-12} 

& RF                                             
& \multicolumn{1}{l|}{5.67}     & 35.31
& \multicolumn{1}{l|}{4.57}     & 28.02
& \multicolumn{1}{l|}{2.66}     & 12.64
& \multicolumn{1}{l|}{2.07}     & 7.58
& \multicolumn{1}{l|}{4.68}     & 24.74  \\ \cline{2-12} 

& ANN                                            
& \multicolumn{1}{l|}{2.29}     & 14.00
& \multicolumn{1}{l|}{\bf 2.05}     & 8.59
& \multicolumn{1}{l|}{0.89}     & 3.85
& \multicolumn{1}{l|}{7.29}     & 24.37
& \multicolumn{1}{l|}{2.47}     & 10.54  \\ \cline{2-12} 

& Ensemble                                       
& \multicolumn{1}{l|}{2.79}     & 14.57
& \multicolumn{1}{l|}{2.67}     & 11.94
& \multicolumn{1}{l|}{1.15}     & 4.35
& \multicolumn{1}{l|}{2.36}     & 10.43
& \multicolumn{1}{l|}{4.07}     & 12.04  \\ \cline{2-12} 

& GCN                                            
& \multicolumn{1}{l|}{\bf 2.16}     & 12.21
& \multicolumn{1}{l|}{2.18}     & 7.77
& \multicolumn{1}{l|}{\bf 0.66}     & 4.02
& \multicolumn{1}{l|}{2.46}     & 6.92
& \multicolumn{1}{l|}{\bf 2.31}     & 8.53  \\ \hline
\multirow{5}{*}{\begin{tabular}[c]{@{}c@{}}Axiline\\ GF12\end{tabular}} 
& GBDT                                           
& \multicolumn{1}{l|}{0.77}     & 5.24
& \multicolumn{1}{l|}{2.20}     & 14.70
& \multicolumn{1}{l|}{2.74}     & 13.59                          
& \multicolumn{1}{l|}{1.34}     & 12.31
& \multicolumn{1}{l|}{1.15}     & 8.87  \\ \cline{2-12} 

& RF                                             
& \multicolumn{1}{l|}{6.55}     & 36.06
& \multicolumn{1}{l|}{3.79}     & 29.70
& \multicolumn{1}{l|}{3.50}     & 16.94                          
& \multicolumn{1}{l|}{\bf 1.32}     & 13.00
& \multicolumn{1}{l|}{7.53}     & 91.34  \\ \cline{2-12} 

& ANN                                            
& \multicolumn{1}{l|}{0.78}     & 8.69
& \multicolumn{1}{l|}{2.78}     & 28.19
& \multicolumn{1}{l|}{2.21}     & 53.32
& \multicolumn{1}{l|}{4.46}     & 77.34
& \multicolumn{1}{l|}{1.29}     & 13.16
\\ \cline{2-12} 

& Ensemble                                       
& \multicolumn{1}{l|}{\bf 0.70}     & 8.50
& \multicolumn{1}{l|}{2.44}         & 28.53
& \multicolumn{1}{l|}{\bf 1.46}     & 20.99
& \multicolumn{1}{l|}{9.15}         & 95.32
& \multicolumn{1}{l|}{\bf 1.05}     & 8.30  \\ \cline{2-12} 

& GCN                                            
& \multicolumn{1}{l|}{3.06}     & 49.65
& \multicolumn{1}{l|}{\bf 1.52} & 22.69
& \multicolumn{1}{l|}{1.82}     & 16.09                          
& \multicolumn{1}{l|}{2.68}     & 37.83
& \multicolumn{1}{l|}{1.39}     & 25.56  \\ \hline
\multirow{5}{*}{\begin{tabular}[c]{@{}c@{}}Axiline\\ NG45\end{tabular}} 
& GBDT                                           
& \multicolumn{1}{l|}{3.56}     & 22.73
& \multicolumn{1}{l|}{7.01}     & 33.61
& \multicolumn{1}{l|}{2.60}     & 12.62
& \multicolumn{1}{l|}{6.43}     & 51.06
& \multicolumn{1}{l|}{\bf 3.95}     & 36.78  \\ \cline{2-12} 

& RF                                             
& \multicolumn{1}{l|}{4.56}     & 30.57
& \multicolumn{1}{l|}{9.38}     & 45.35
& \multicolumn{1}{l|}{3.70}     & 13.38
& \multicolumn{1}{l|}{6.92}     & 41.56
& \multicolumn{1}{l|}{4.21}     & 44.36  \\ \cline{2-12} 

& ANN                                            
& \multicolumn{1}{l|}{3.48}  & 25.40
& \multicolumn{1}{l|}{8.48}     & 83.22
& \multicolumn{1}{l|}{1.93}      & 25.85
& \multicolumn{1}{l|}{7.04}      & 51.25
& \multicolumn{1}{l|}{6.60}      & 45.50  \\ \cline{2-12} 
                                                                        
& Ensemble                                       
& \multicolumn{1}{l|}{\bf 3.15}     & 23.61
& \multicolumn{1}{l|}{7.68}     & 54.21
& \multicolumn{1}{l|}{\bf 1.39}     & 8.81
& \multicolumn{1}{l|}{8.91}     & 75.83
& \multicolumn{1}{l|}{5.16}     & 31.31  \\ \cline{2-12} 

& GCN                                            
& \multicolumn{1}{l|}{4.74}     & 36.25
& \multicolumn{1}{l|}{\bf 5.19}    & 29.98
& \multicolumn{1}{l|}{3.03}     & 13.48
& \multicolumn{1}{l|}{\bf 4.97}    & 25.07
& \multicolumn{1}{l|}{4.59}     & 55.06  \\ \hline
\end{tabular}
\end{table}

\smallskip
\noindent
{\bf Results for unseen architectural configurations.}
\sk{
Table~\ref{tab:unseen_arch} shows the performance
of the ML model for unseen architectural configurations in
predicting post-SP\&R PPA and system-level metrics for
TABLA, Genesys, VTA, and Axiline designs, implemented
on GF12 and/or NG45. For the ROI classification task
all the models for GF12 implementation achieve at 
least 95\% accuracy and 0.97 F1 score. This indicates
that models are performing very well in 
determining whether the data points belong to the ROI. 
In Table~\ref{tab:unseen_arch}, the best-performing
model based on $\mu APE$ for each design and each metric
is highlighted in bold. Based on this table and for
these models, we make the following observations.}
\sk{
\begin{itemize}[noitemsep,topsep=0pt,leftmargin=*]
    \item The best-performing ML model achieves a 
    $\mu APE$ of less than 7\% and a $MAPE$
    of less than 37\% for backend PPA prediction.
    For Axiline-NG45, the backend power 
    prediction using the GCN model yields the highest
    $MAPE$. Upon examining the model, we find that
    the STD $APE$ on the testing dataset is
    4.49.
    \item The best-performing ML model achieves 
    a $\mu APE$ of less than 8\% and a 
    $MAPE$ of less than 50\% for system-level
    metric prediction. 
    For Axiline-NG45, the system-level energy
    prediction using the Ensemble model yields
    the highest $MAPE$. Upon investigating the
    model, we find that the STD $APE$ on the
    testing dataset is 7.07.
\end{itemize}
}

\begin{table}[htbp]
\centering
\caption{Performance of ML models for unseen architectural configurations.}
\label{tab:unseen_arch}
\customsize
\begin{tabular}{|c|l|ll|ll|ll|ll|ll|}
\hline
\multirow{2}{*}{Design}
& \multicolumn{1}{c|}{\multirow{2}{*}{ML Model}}
& \multicolumn{2}{c|}{Backend-Perf}
& \multicolumn{2}{c|}{Backend-Power}
& \multicolumn{2}{c|}{Backend-Area}
& \multicolumn{2}{c|}{System-Energy}
& \multicolumn{2}{c|}{System-Runtime} \\ \cline{3-12}
    & \multicolumn{1}{c|}{}
    & \multicolumn{1}{c|}{$\mu APE$} & \multicolumn{1}{c|}{$MAPE$}
    & \multicolumn{1}{c|}{$\mu APE$} & \multicolumn{1}{c|}{$MAPE$} 
    & \multicolumn{1}{c|}{$\mu APE$} & \multicolumn{1}{c|}{$MAPE$} 
    & \multicolumn{1}{c|}{$\mu APE$} & \multicolumn{1}{c|}{$MAPE$} 
    & \multicolumn{1}{c|}{$\mu APE$} & \multicolumn{1}{c|}{$MAPE$} \\ \hline
\multirow{5}{*}{\begin{tabular}[c]{@{}c@{}}TABLA\\ GF12\end{tabular}}   
& GBDT
& \multicolumn{1}{l|}{\bf 3.24}      & 33.06
& \multicolumn{1}{l|}{\bf 3.87}      & 19.03
& \multicolumn{1}{l|}{3.42}      & 9.01
& \multicolumn{1}{l|}{8.86}      & 19.15
& \multicolumn{1}{l|}{\bf 3.83}      & 18.99 \\ \cline{2-12} 

& RF                                             
& \multicolumn{1}{l|}{5.16}     & 38.01
& \multicolumn{1}{l|}{9.76}     & 46.62
& \multicolumn{1}{l|}{11.25}     & 40.13
& \multicolumn{1}{l|}{10.59}     & 21.30
& \multicolumn{1}{l|}{5.67}     & 27.41  \\ \cline{2-12} 

& ANN                                            
& \multicolumn{1}{l|}{5.78}     & 32.70
& \multicolumn{1}{l|}{5.22}     & 20.02
& \multicolumn{1}{l|}{\bf 2.30}     & 5.70
& \multicolumn{1}{l|}{\bf 2.97}     & 10.25
& \multicolumn{1}{l|}{6.02}     & 24.94  \\ \cline{2-12} 

& Ensemble                                       
& \multicolumn{1}{l|}{3.68}     & 32.51
& \multicolumn{1}{l|}{4.11}     & 17.11
& \multicolumn{1}{l|}{3.99}     & 16.05
& \multicolumn{1}{l|}{4.62}     & 18.63
& \multicolumn{1}{l|}{6.03}     & 24.10  \\ \cline{2-12}

& GCN
& \multicolumn{1}{l|}{5.79}     & 21.74
& \multicolumn{1}{l|}{5.34}     & 14.00
& \multicolumn{1}{l|}{3.76}     & 12.81
& \multicolumn{1}{l|}{3.93}     & 13.80
& \multicolumn{1}{l|}{5.20}     & 23.63  \\ \hline

\multirow{5}{*}{\begin{tabular}[c]{@{}c@{}}GeneSys\\GF12\end{tabular}} 
& GBDT                                           
& \multicolumn{1}{l|}{6.54}     & 20.86
& \multicolumn{1}{l|}{5.82}     & 15.99
& \multicolumn{1}{l|}{2.94}     & 8.52
& \multicolumn{1}{l|}{6.37}     & 11.65
& \multicolumn{1}{l|}{12.23}     & 28.41  \\ \cline{2-12}

& RF                                             
& \multicolumn{1}{l|}{8.73}     & 22.57
& \multicolumn{1}{l|}{9.34}     & 18.66
& \multicolumn{1}{l|}{3.69}     & 9.57
& \multicolumn{1}{l|}{14.89}     & 22.00
& \multicolumn{1}{l|}{17.37}     & 34.38 \\ \cline{2-12} 

& ANN                                            
& \multicolumn{1}{l|}{6.55}     & 19.86
& \multicolumn{1}{l|}{\bf 3.82}      & 15.36
& \multicolumn{1}{l|}{\bf 2.06}     & 3.80
& \multicolumn{1}{l|}{\bf 3.47}      & 9.11
& \multicolumn{1}{l|}{10.73}     & 28.90 \\ \cline{2-12} 

& Ensemble   
& \multicolumn{1}{l|}{\bf 6.32}     & 14.82
& \multicolumn{1}{l|}{7.26}     & 15.23
& \multicolumn{1}{l|}{2.75}     & 8.09
& \multicolumn{1}{l|}{11.96}    & 19.78
& \multicolumn{1}{l|}{\bf 6.28}     & 20.27 \\ \cline{2-12} 

& GCN                                            
& \multicolumn{1}{l|}{6.97}    & 13.11
& \multicolumn{1}{l|}{5.39}     & 15.10
& \multicolumn{1}{l|}{2.12}     & 4.08
& \multicolumn{1}{l|}{4.32}     & 8.88
& \multicolumn{1}{l|}{7.65}     & 17.81  \\ \hline
\multirow{5}{*}{\begin{tabular}[c]{@{}c@{}}VTA\\ GF12\end{tabular}}     
& GBDT                                           
& \multicolumn{1}{l|}{4.96}     & 19.31
& \multicolumn{1}{l|}{4.04}     & 11.83
& \multicolumn{1}{l|}{24.74}     & {38.07}
& \multicolumn{1}{l|}{7.58}     & 18.79
& \multicolumn{1}{l|}{6.94}     & 20.39  \\ \cline{2-12} 

& RF                                             
& \multicolumn{1}{l|}{3.00}     & 9.81
& \multicolumn{1}{l|}{12.96}    & 33.34
& \multicolumn{1}{l|}{18.05}    & 53.58
& \multicolumn{1}{l|}{7.15}     & 16.43
& \multicolumn{1}{l|}{5.67}     & 13.52  \\ \cline{2-12} 

& ANN                                            
& \multicolumn{1}{l|}{\bf 2.52}     & 14.09
& \multicolumn{1}{l|}{3.08}     & 11.84
& \multicolumn{1}{l|}{2.19}     & 6.66
& \multicolumn{1}{l|}{10.61}     & 22.16
& \multicolumn{1}{l|}{4.39}     & 12.70  \\ \cline{2-12} 

& Ensemble                                       
& \multicolumn{1}{l|}{2.99}     & 12.58
& \multicolumn{1}{l|}{11.19}     & 28.99
& \multicolumn{1}{l|}{6.65}     & 17.01
& \multicolumn{1}{l|}{9.10}     & 18.41
& \multicolumn{1}{l|}{\bf 2.87}     & 10.50  \\ \cline{2-12} 

& GCN
& \multicolumn{1}{l|}{2.60}     & 9.67
& \multicolumn{1}{l|}{\bf 2.85}     & 12.90
& \multicolumn{1}{l|}{\bf 2.15}     & 9.51
& \multicolumn{1}{l|}{\bf 4.07}     & 13.76
& \multicolumn{1}{l|}{3.67}     & 9.87  \\ \hline
\multirow{5}{*}{\begin{tabular}[c]{@{}c@{}}Axiline\\ GF12\end{tabular}} 
& GBDT
& \multicolumn{1}{l|}{0.62}      & 7.18
& \multicolumn{1}{l|}{11.53}     & 74.19
& \multicolumn{1}{l|}{10.29}     & 41.78
& \multicolumn{1}{l|}{16.61}     & 82.95
& \multicolumn{1}{l|}{2.19}      & 19.83  \\ \cline{2-12} 

& RF
& \multicolumn{1}{l|}{0.63}     & 5.41
& \multicolumn{1}{l|}{15.95}    & 77.38
& \multicolumn{1}{l|}{13.24}    & 57.18
& \multicolumn{1}{l|}{21.8}     & 90.34
& \multicolumn{1}{l|}{2.12}     & 12.86  \\ \cline{2-12} 

& ANN
& \multicolumn{1}{l|}{0.72}     & 8.64
& \multicolumn{1}{l|}{\bf 2.24}     & 21.98
& \multicolumn{1}{l|}{\bf 1.20}     & 7.85
& \multicolumn{1}{l|}{4.24}     & 29.85
& \multicolumn{1}{l|}{\bf 1.08}     & 9.72  \\ \cline{2-12} 

& Ensemble
& \multicolumn{1}{l|}{\bf 0.61} & 6.48
& \multicolumn{1}{l|}{2.55}     & 22.45
& \multicolumn{1}{l|}{1.31}     & 5.68
& \multicolumn{1}{l|}{7.17}     & 47.20
& \multicolumn{1}{l|}{1.29}     & 7.74  \\ \cline{2-12} 

& GCN
& \multicolumn{1}{l|}{2.92}     & 28.74
& \multicolumn{1}{l|}{2.86}     & 29.34
& \multicolumn{1}{l|}{1.88}     & 9.82
& \multicolumn{1}{l|}{\bf 2.34}     & 29.21
& \multicolumn{1}{l|}{2.98}     & 2971  \\ \hline
\multirow{5}{*}{\begin{tabular}[c]{@{}c@{}}Axiline\\ NG45\end{tabular}} 
& GBDT
& \multicolumn{1}{l|}{3.45}     & 27.48
& \multicolumn{1}{l|}{5.98}     & 56.62
& \multicolumn{1}{l|}{2.79}     & 13.40
& \multicolumn{1}{l|}{23.33}    & 90.79
& \multicolumn{1}{l|}{5.54}     & 30.13  \\ \cline{2-12} 

& RF
& \multicolumn{1}{l|}{\bf 3.18}     & 26.96
& \multicolumn{1}{l|}{6.30}     & 41.31
& \multicolumn{1}{l|}{3.00}     & 16.90
& \multicolumn{1}{l|}{21.54}    & 93.74
& \multicolumn{1}{l|}{5.77}     & 35.33  \\ \cline{2-12}

& ANN                                            
& \multicolumn{1}{l|}{3.37}     & 27.19
& \multicolumn{1}{l|}{6.57}     & 59.76
& \multicolumn{1}{l|}{\bf 1.86}     & 13.81
& \multicolumn{1}{l|}{9.30}    & 77.10
& \multicolumn{1}{l|}{5.04}     & 25.56  \\ \cline{2-12} 

& Ensemble                                       
& \multicolumn{1}{l|}{3.33}     & 27.29
& \multicolumn{1}{l|}{5.97}     & 48.34
& \multicolumn{1}{l|}{2.81}     & 17.25
& \multicolumn{1}{l|}{\bf 7.21}     & 49.81
& \multicolumn{1}{l|}{\bf 4.75}     & 22.63  \\ \cline{2-12} 

& GCN                                            
& \multicolumn{1}{l|}{4.57}     & 35.68
& \multicolumn{1}{l|}{\bf 5.55}     & 36.38
& \multicolumn{1}{l|}{3.77}     & 16.26
& \multicolumn{1}{l|}{12.88}    & 86.2
& \multicolumn{1}{l|}{5.85}     & 51.21  \\ \hline
\end{tabular}
\end{table}

\sk{For some unseen architectural configurations,
we observe MAPE values higher than 30\%. However,
further investigation reveals smaller standard
deviation values for the $\mu APE$. This
indicates that our model delivers reliable
results for most of the data points.
We also observe that for both unseen backend and
unseen
architectural configuration datasets, the GCN model
outperforms other models in most scenarios.
In instances where it
does not come out on top, it still yields results very
similar to those of the best-performing model.}

\subsection{Effect of Limited Training Dataset}
\label{subsec:extrapolation}
\begin{figure}
    \centering
    \includegraphics[width=0.49\columnwidth]{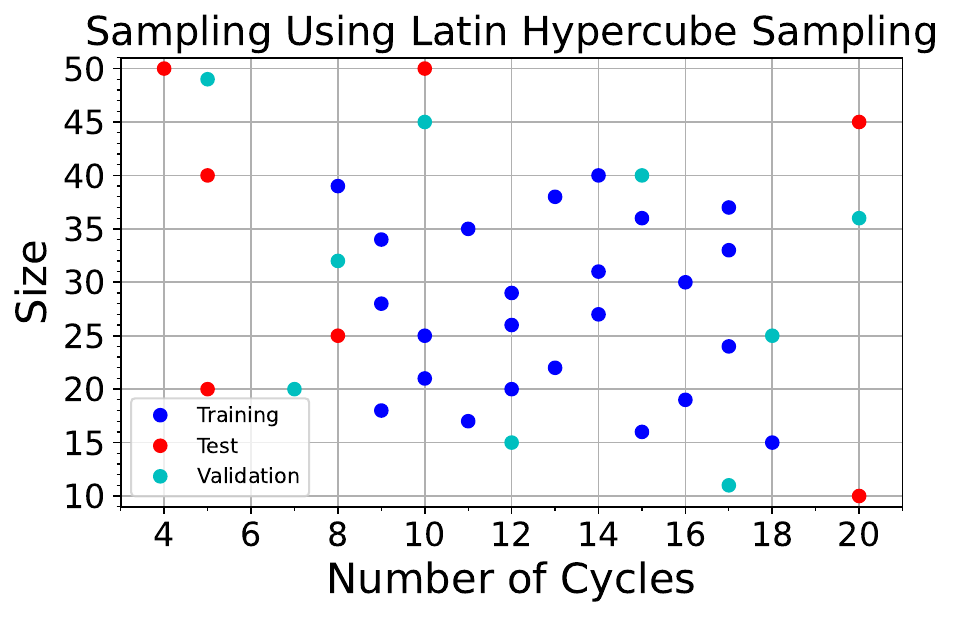}
    \caption{\sk{Sampled train, validation and test
    data points for extrapolation experiment.}}
    \label{fig:sample_extrapolation}
\end{figure}

\sk{
We have additionally studied the performance of our model
in terms of $extrapolation$. In other words we study how
well our model performs when the test data points fall 
outside the range of the training dataset. For this
experiment, we sample architectural configurations of the
Axiline design as displayed in 
Figure~\ref{fig:sample_extrapolation}, where the blue,
red, and cyan data points correspond to training, testing,
and validation data respectively. For each architectural
configuration, we run 30 SP\&R jobs to prepare the training, 
testing, and validation datasets. We observe that the model 
performs poorly on the validation dataset and produces similar
poor results for the testing dataset, confirming the 
limitations of the ML model we use for dataset extrapolation.}

\sk{However, our training dataset covers the 
architectural design space for Axiline and backend design
space for all accelerators including GeneSys, VTA, TABLA and
Axiline. The count of features handled by the Axiline design
is computed as $num\_cycles \times size$. The Axiline designs
are expected to handle up to 800 features~\cite{VeriGOOD-ML}
and our
sample space already covers up to 900 features. As seen in
Figure~\ref{fig:axiline_ecp_vs_tcp}, the $\ecf$ does not change
for higher $\tcf$, indicating that we have effectively covered
the backend design space. Therefore, even though the models
fail to produce satisfactory results for the extrapolation
dataset, it is not considered a drawback. Our training's
configuration space includes the entire design space,
eliminating the need for the model to function outside the
scope of the training configurations.}

\subsection{DSE with Trained ML Models}
\label{subsec:dse_application}
\sk{We now apply our trained models for DSE on two different
platforms: Axiline and VTA.
We apply the MOTPE method and our trained models
for the DSE of the ML accelerator,
with the objective of minimizing chip area and system
energy. We also employ an additional flag during the
DSE process to indicate whether the explored data points
fulfill design configuration, runtime, and power
requirements. The specifics of these two DSE experiments
are as follows.}
  
\begin{figure}
    \centering
    \begin{subfigure}[]
        {\includegraphics[width=0.45\columnwidth]{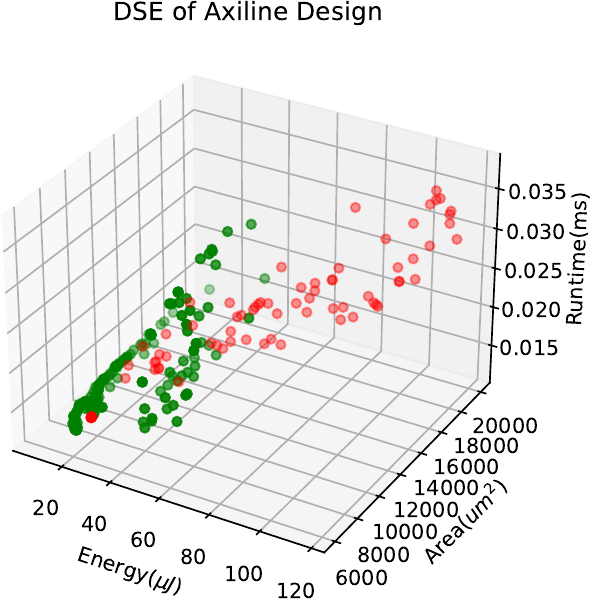}}
    \end{subfigure}
    \begin{subfigure}[]
        {\includegraphics[width=0.45\columnwidth]{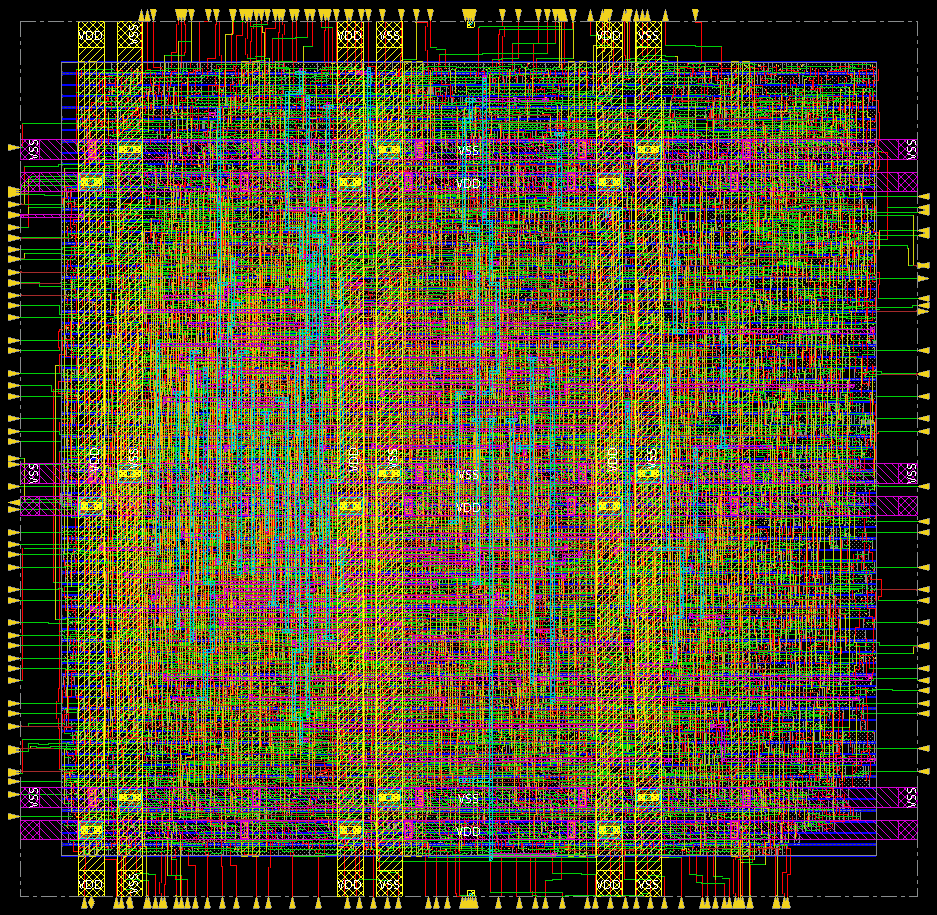}}
    \end{subfigure}
    \caption{\sk{Design space exploration of the Axiline-SVM designs. 
    (a) Energy, runtime, and area metrics of the explored 
    data points. Data points highlighted in red do not 
    meet the ROI, power and runtime criteria. (b) Layout of the SVM design
    with the lowest cost based on chip area and system-level
    energy of the green dots.}}
    \label{fig:dse_svm}
\end{figure}

\noindent
\sk{{\bf DSE of Axiline-SVM in NG45 enablement.} We
optimize the implementation of an 
accelerator that executes the support vector machine 
(SVM) algorithm with 55 features. The flow is as follows.
\begin{itemize}[noitemsep,topsep=0pt,leftmargin=*]
\item We select a range for architectural and backend
parameter configurations.
\item For the given set of configurations, we carry out
DSE using MOTPE and trained models, capturing energy,
runtime, total power, and chip area for the sampled
configurations.
\item We identify the best configuration that minimizes
Equation~\eqref{eq:DSE} when $\alpha$ is 1 and $\beta$
is 0.001.
\item For the best configuration, we generate the RTL netlist, 
run SP\&R and collect the actual backend and chip parameters.
\end{itemize}
}
\sk{
\noindent
During the DSE process, we vary $size$ from 10 to 51,
$num\_cycle$ from 5 to 21, $\tcf$ from 0.3 to 1.3, 
and floorplan utilization from 0.4 to 0.8.
Figure~\ref{fig:dse_svm}(a) displays the predicted
runtime, area, and energy plot for all the data points
sampled during the DSE process. The data points
highlighted with red dots do not meet the ROI, power
and runtime requirements. We then identify the best
configuration that minimizes Equation~\eqref{eq:DSE}.
Figure~\ref{fig:dse_svm}(b) presents the chip layout
for the best configuration obtained during DSE.
For ground truth analysis, we also generate
the RTL netlist and run SP\&R for each of
the top three configurations, and
confirm that the predicted metrics for these
top three configurations are within 7\% of 
post-SP\&R values.
}

\begin{figure}
    \centering
    \begin{subfigure}[]
        {\includegraphics[width=0.45\columnwidth]{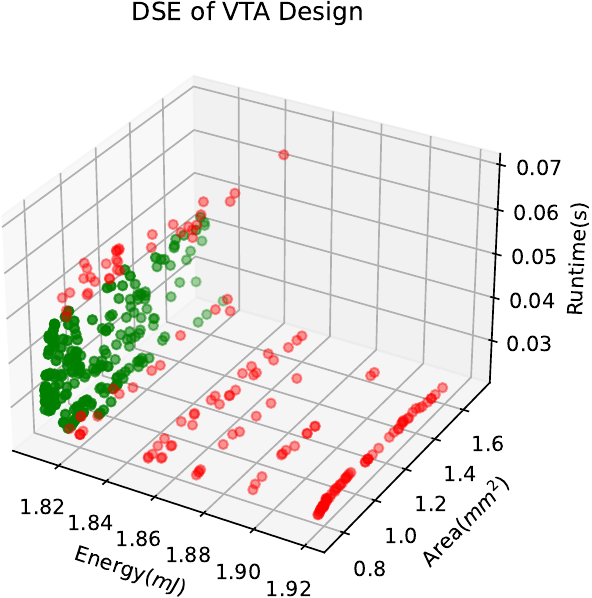}}
    \end{subfigure}
    \begin{subfigure}[]
        {\includegraphics[width=0.45\columnwidth]{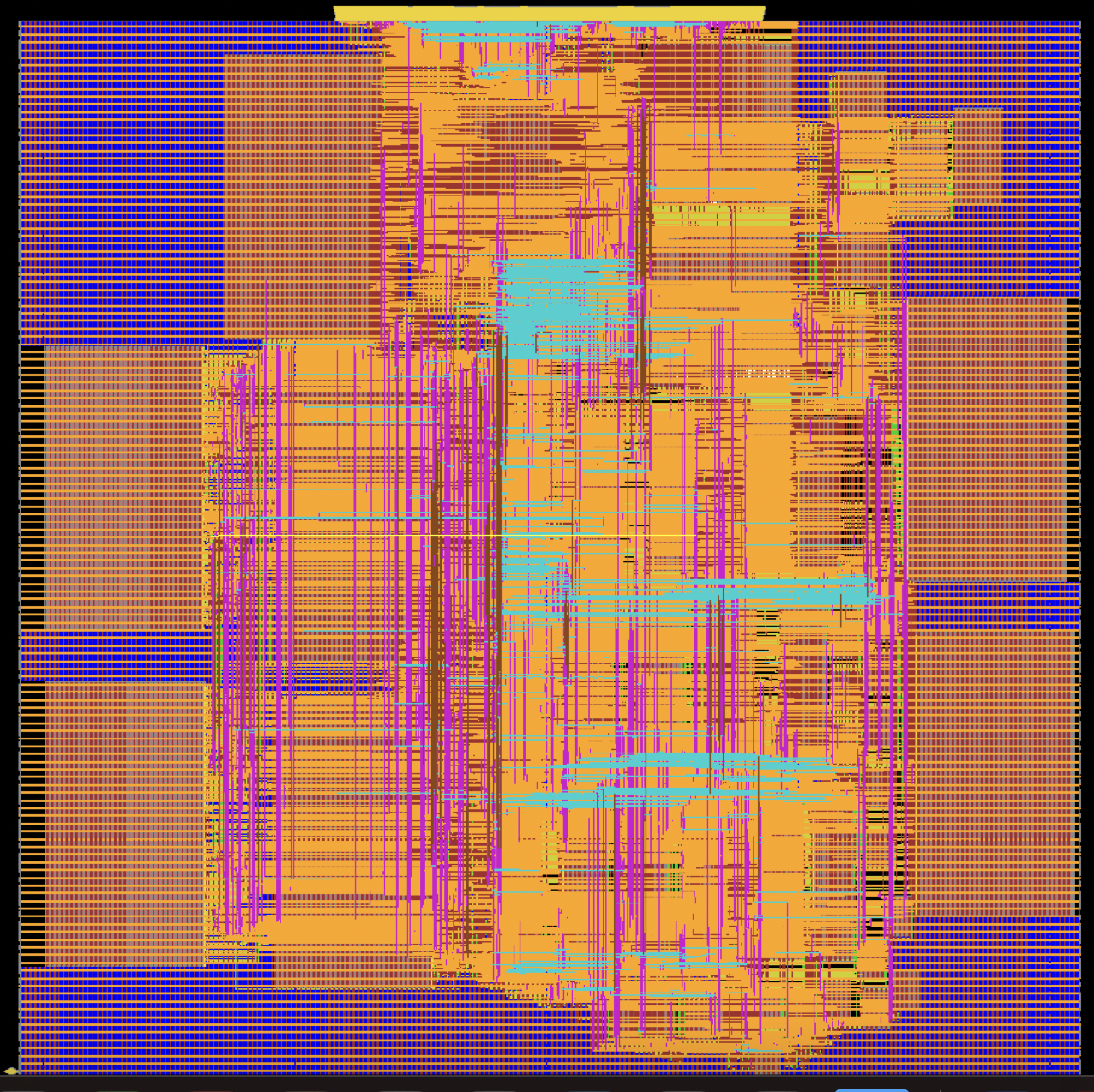}}
    \end{subfigure}
    \caption{\sk{Backend design space exploration of a VTA design. 
    (a) Energy, runtime, and area metrics of the explored 
    data points. Data points highlighted in red do not 
    meet the ROI, power and runtime criteria.
    (b) Layout of the VTA design
    with the lowest cost based on chip area, system-level
    energy of the green dots.}}
    \label{fig:dse_vta}
\end{figure}

\sk{
\noindent
{\bf DSE of VTA design in GF12 enablement.} We also apply our
the DSE method 
to optimize the backend configuration for the VTA design. The
process is the same as outlined above for Axiline-SVM, except we 
only vary $\tcf$ from 0.3 to 1.3 and floorplan utilization from
0.25 to 0.55. Figure~\ref{fig:dse_vta}(a) displays the predicted
runtime, area, and energy plots for all the data points sampled
during the DSE process. The data points marked with red dots do
not meet the ROI, power, and runtime requirements. We identify
the best configuration that minimizes Equation \eqref{eq:DSE} when
$\alpha$ and $\beta$ are both 1. We adjust these values because the
units of energy and runtime in Figure~\ref{fig:dse_vta}(a) differ
from those in Figure~\ref{fig:dse_svm}(a). Figure~\ref{fig:dse_vta}(b) 
displays the chip layout for the best configuration obtained during
the DSE. Running SP\&R on the top three backend configurations 
identified from the DSE confirms that 
the predicted metrics for these top three configurations are within
6\% of the post-SP\&R values.
}

\section{Conclusion}
\label{sec:conclusion}
\sk{We introduce a physical-design-driven, ML-based framework that
consistently predicts backend PPA and system metrics with an
average 7\% or less prediction error for the ASIC implementation
of two deep learning accelerator platforms: VTA and VeriGOOD-ML,
in both a commercial 12 nm process and a research-oriented 45 nm process.
Our work encompasses an extensive study of the performance
of various sampling methods.} The framework integrates several
ML modeling aspects, including a focus on the ``region of interest",
a novel two-stage approach, and \sk{the employment of logical
hierarchy graphs for a GCN model, enabling efficient
model-guided MOTPE-based automated searches over vast
accelerator architecture and backend configuration spaces
for a given workload or ML algorithm.} We extensively
validate our framework on multiple ML accelerator platforms.

\begin{acks}
This material is based on research sponsored in part by Air Force 
Research Laboratory (AFRL) and Defense Advanced Research Projects Agency
(DARPA) under agreement number FA8650-20-2-7009. Andrew B. Kahng also
acknowledges support from NSF CCF-2112665. The U. S.
government is authorized to reproduce and distribute reprints for
Governmental purposes notwithstanding any copyright notation thereon.
The views and conclusions
contained herein are those of the authors and should not be interpreted as
necessarily representing the official policies or endorsements, either
expressed or implied, of AFRL, DARPA, or the U. S. government.

The authors
would like to acknowledge the contributions of Steven M. Burns
and Anton A. Sorokin from Intel Labs.
\end{acks}

\end{document}